%% file: main.tex
\pdfoutput=1
\documentclass[11pt]{article}

\PassOptionsToPackage{dvipsnames}{xcolor}
\usepackage[final]{acl}

\usepackage{times}
\usepackage{latexsym}

\usepackage{microtype}

\usepackage{inconsolata}

\usepackage{graphicx}

\usepackage[utf8]{inputenc} 
\usepackage[T1]{fontenc}    
\usepackage{hyperref}       
\usepackage{url}            
\usepackage{booktabs}       
\usepackage{amsfonts}       
\usepackage{nicefrac}       
\usepackage{xcolor} 
\usepackage{todonotes}
\usepackage{enumitem}
\usepackage{amsmath}

\usepackage{bbm}
\usepackage[most]{tcolorbox}    
\usepackage{mdframed}
\definecolor{LightGray}{rgb}{0.9, 0.9, 0.9}

\newtcolorbox{mybox}[1]{colback=red!5!white,colframe=red!75!black,fonttitle=\bfseries,title=#1}

\newcommand{\ME}[1]{#1}

\newcommand{\CRPO}{\textsc{CrPO}}
\newcommand{\MUSE}{\textsc{MuCE}}

\title{Creative Preference Optimization}

\author{
 \textbf{Mete Ismayilzada\textsuperscript{1,2}},
 \textbf{Antonio Laverghetta Jr.\textsuperscript{*3}},
 \textbf{Simone A. Luchini\textsuperscript{4}}, \\
 \textbf{Reet Patel\textsuperscript{4}},
 \textbf{Antoine Bosselut\textsuperscript{1}},
 \textbf{Lonneke van der Plas \textsuperscript{\textdagger2}},
 \textbf{Roger E. Beaty \textsuperscript{\textdagger4}}
\\
\\
 \textsuperscript{1}EPFL,
 \textsuperscript{2}Università della Svizzera Italiana,
 \textsuperscript{3}Wesleyan University, \\
 \textsuperscript{4}Pennsylvania State University \\
 \small{
   \href{mailto:email@domain}{mahammad.ismayilzada@epfl.ch}
 }
}

\begin{document}

\maketitle

\begingroup
\renewcommand\thefootnote{\textsuperscript{*}}\footnotetext{Work done while at Pennsylvania State University}
\endgroup

\begingroup
\renewcommand\thefootnote{\textsuperscript{\textdagger}}\footnotetext{Equal supervision}
\endgroup

\begin{abstract}
\input{00_abstract}
\end{abstract}

\section{Introduction}
\input{01_introduction}

\section{Related Work}
\input{02_related_work}

\section{Creative Preference Optimization}
\label{sec:cdpo}
\input{03_cdpo}

\section{The \MUSE{} Dataset}
\input{04_MUSE}

\section{Experiments}
\input{05_experiments}

\section{Results}

\input{06_results}

\section{Discussion}
\input{07_discussion}

\section{Conclusion}

\input{08_conclusion}

\input{09_limitations}

\section*{Acknowledgements}
\input{acknowledgements}

\bibliography{literature}

\appendix
\input{10_appendix}

\end{document}

%% file: 00_abstract.tex

While Large Language Models (LLMs) have demonstrated impressive performance across natural language generation tasks, their ability to generate truly creative content—characterized by novelty, diversity, surprise, and quality—remains limited. Existing methods for enhancing LLM creativity often focus narrowly on diversity or specific tasks, failing to address creativity’s multifaceted nature in a generalizable way. In this work, we propose Creative Preference Optimization (\textsc{CrPO}), a novel alignment method that injects signals from multiple creativity dimensions into the preference optimization objective in a modular fashion. We train and evaluate creativity-augmented versions of several models using \textsc{CrPO} and \textsc{MuCE}, a new large-scale human preference dataset spanning over 200,000 human-generated responses and ratings from more than 30 psychological creativity assessments. Our models outperform strong baselines, including GPT-4o, on both automated and human evaluations, producing more novel, diverse, and surprising generations while maintaining high output quality. Additional evaluations on \textsc{NoveltyBench} further confirm the generalizability of our approach. Together, our results demonstrate that directly optimizing for creativity within preference frameworks is a promising direction for advancing the creative capabilities of LLMs without compromising output quality.


%% file: 01_introduction.tex

\begin{figure*}[t]
    \centering   
    \includegraphics[width=\linewidth,trim={0cm 4.5cm 0cm 0cm},clip]{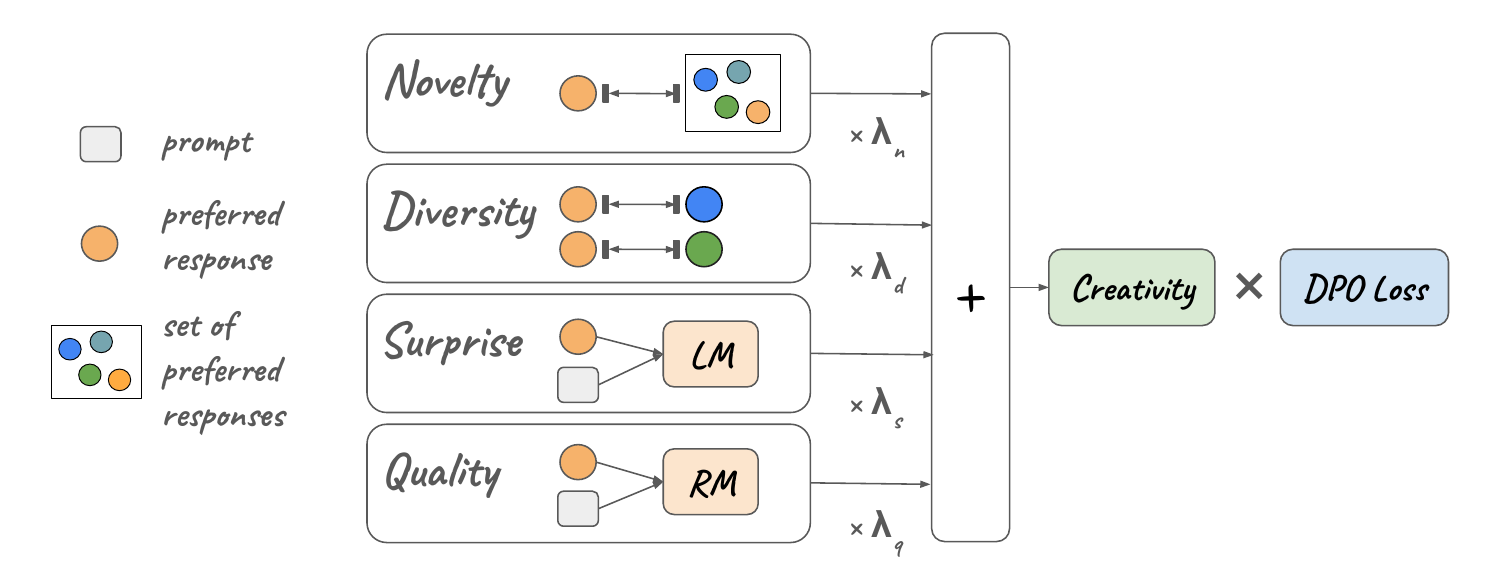}
    \caption{Our preference alignment method \CRPO{} to improve output creativity by injecting a weighted combination of signals from multiple creativity dimensions.}
    \label{fig:main-figure}
\end{figure*} 

Large Language Models (LLMs) have made significant progress across a broad range of natural language generation tasks \cite{team2023gemini, zhao2025surveylargelanguagemodels, bubeck2023sparks, wei2022emergent, brown2020language}. However, whether LLMs exhibit true human-like creativity, i.e, the ability to produce novel (i.e., original), high-quality (i.e., useful) and surprising (i.e, unexpected) ideas \cite{simonton2012taking, boden2004creative} remains unclear. Research on the creativity of LLMs has found mixed results, with some reporting that LLMs are more creative than humans \cite{bellemare2024divergent, zhao2024assessing, stevenson2022putting}, others reporting that they are less creative \cite{koivisto2023best, chakrabarty2024art, ismayilzada2024evaluating}, and some finding their creativity to be on par with each other \cite{GesPushingGC, gilhooly2024ai}. However, past research has also found that the high LLM performance can be attributed to the artificial nature of the creativity tasks \cite{ismayilzada2024creativity} commonly employed to evaluate LLMs, such as the Alternative Uses Task \cite{guilford1967nature} or to the remarkable creativity of human-written texts on the web \cite{lu2024ai}. Consequently, LLMs have been shown to often lack novelty and surprise in their generations \cite{ismayilzada2024creativity, ismayilzada2024evaluating, zhang2025noveltybench, tian2024large, chakrabarty2024art} and produce significantly less diverse content compared to humans \cite{padmakumar2023does, anderson2024homogenization, kirk2023understanding, xu2024echoes, o2024attributing, zhang2024forcing, wenger2025we}. These tendencies limit the utility of LLMs for creative tasks, such as story generation and creative problem solving that often require longer responses and ``out-of-the-box'' thinking \cite{tian2023macgyver,huang-etal-2024-lateval,chen2024weak}.

Recent research has proposed some methods for improving the creativity of LLMs, often targeting the diversity aspect alone \cite{wong2024simplestrat, hayati2023far, chung2023increasing, franceschelli2024creative, zhang2024forcing, wang2024weaver, zhou2025bridging, lanchantin2025diverse, chung2025modifying} or focusing on a single creativity task \cite{tian2023macgyver, nair-etal-2024-creative, SummersStay2023BrainstormTS}. However, creativity is a multifaceted ability that also encompasses novelty, surprise, and quality and manifests itself in a wide range of tasks. Consequently, it has been argued that methods promoting creativity improvements should consider multiple dimensions of creativity together across several creative tasks \cite{ismayilzada2024creativity}. Hence, the broader challenge of enhancing overall creativity in LLM outputs largely remains underexplored.

To this end, we propose a novel approach to directly optimize for creativity in language model generation through preference learning \cite{ouyang2022training, rafailov2023direct}. Recent works targeting improvement in LLM creativity have mainly focused on black-box techniques to elicit creative outputs through input-level (e.g., prompting) \cite{tian2023macgyver, mehrotra2024enhancing, nair-etal-2024-creative, SummersStay2023BrainstormTS} and output-level strategies (e.g., creative decoding) \cite{nguyen2024turning, franceschelli2024creative, meister-etal-2023-locally}. However, these methods are inherently limited to the fixed creative capacity of language models and are not designed to optimize for fine-grained dimensions of creativity. Recently, motivated by the negative impact of the preference alignment techniques on the diversity of LLM outputs \cite{padmakumar2023does, anderson2024homogenization, kirk2023understanding, o2024attributing, west2025basemodelsbeataligned}, few works have suggested directly modifying the preference optimization methods to promote output diversity \cite{lanchantin2025diverse, chung2025modifying}. Inspired by these approaches, we design a new optimization strategy that \textit{injects} signals from multiple dimensions of creativity into the preference modeling objective in a \textit{modular} fashion. Specifically, we integrate the novelty, diversity, surprise, and quality dimensions of creativity into the training objective of direct preference optimization (DPO) \cite{rafailov2023direct}, with weighted composition that allows balancing each dimension's contribution. We call this method \textit{creative preference optimization} (\CRPO{}) and provide its conceptual illustration in Figure \ref{fig:main-figure} with full details in Section \ref{sec:cdpo}. 

We test the efficacy of \CRPO{} using \MUSE{} (\textbf{Mu}ltitask \textbf{C}reativity \textbf{E}valuation), our newly curated large-scale dataset of prompt-response pairs annotated with human preferences across a diverse range of creative tasks in multiple languages. While previous work has largely evaluated creativity improvements on a narrow range of tasks like story generation \cite{ismayilzada2024evaluating, chung2025modifying, lanchantin2025diverse} or creative problem solving \cite{tian2023macgyver}, \MUSE{} enables us to test whether our methods truly generalize across a diverse range of creativity assessments.
Our results show that \texttt{Llama-3.1-8B-Instruct} \cite{llama3modelcard} and \texttt{Mistral-7B-Instruct-v0.3} \cite{jiang2023mistral7b} trained using  \CRPO{} outperform the same models trained using only supervised fine-tuning (SFT) or DPO without any creativity injections, as well as existing LLMs such as GPT-4o, generating more novel, diverse, and surprising outputs than all the baselines while maintaining high quality. We publicly release our code, models, and data for future research\footnote{https://www.mete.is/creative-preference-optimization/}.

\textbf{Our main contributions are as follows:} 
\begin{enumerate}
    \item We introduce \MUSE{}, a large-scale preference dataset consisting of more than 200,000 human responses and ratings for more than 30 creativity assessments. All tasks within \MUSE{} are carefully chosen to provide valid measures of creativity in humans, making \MUSE{} one of the largest psychologically valid datasets of human creativity for training preference models.
    \item We propose a novel flexible preference alignment method \CRPO{} that injects signals from several dimensions of creativity into the existing preference optimization method DPO and train creativity-enhanced versions of \texttt{Llama-3.1-8B-Instruct} and \texttt{Mistral-7B-Instruct-v0.3}.
    \item We evaluate the effectiveness of our approach on a range of creativity tasks from \MUSE{}, as well as external tasks from \textsc{NoveltyBench} \cite{zhang2025noveltybench}, using both automated metrics and human evaluations. Our analysis shows that \CRPO{} is a promising method for enhancing the creative capabilities of language models while maintaining quality.
\end{enumerate}

%% file: 02_related_work.tex
\subsection{Large Language Model Creativity}
The potential of building LLM applications for creative industries has spurred significant research interest on AI creativity \cite{bellemare2024divergent}, and many LLM tools marketed for assistance with creative tasks have been developed in the last few years \cite{wang2024weaver}. Yet debates on whether AI is capable of true creativity are nearly as old as AI itself \cite{stein2014stimulating,franceschelli2024creative,saebo2024stochastics}, with theoretical and philosophical arguments being made both for and against AI creativity \cite{ismayilzada2024creativity}. Classic psychological theories of creativity generally agree that, for a product to be creative, it must be new, surprising, and valuable \cite{boden2004creative}. Creative tasks are also often characterized by high diversity \cite{padmakumar2023does, shypula2025evaluating}, though diversity is only one facet of creativity \cite{johnson2021neglect}. Studies on LLM creativity have yielded conflicting findings: some suggest LLMs surpass human creativity \cite{bellemare2024divergent, zhao2024assessing}, others argue they fall short \cite{koivisto2023best, chakrabarty2024art, ismayilzada2024evaluating}, while some conclude that LLM and human creativity are roughly equivalent \cite{gilhooly2024ai, stevenson2022putting, GesPushingGC}. Some works have suggested that LLMs lack novelty and surprise in their generations \cite{ismayilzada2024creativity, ismayilzada2024evaluating, zhang2025noveltybench, tian2024large, chakrabarty2024art} and their seemingly remarkable creative outputs may be in large part attributable to the remarkable creativity of human-written texts on the web \cite{lu2024ai}.
Some recent works have suggested improving the creativity of LLMs through prompting techniques \cite{tian2023macgyver, mehrotra2024enhancing, nair-etal-2024-creative, SummersStay2023BrainstormTS} and decoding strategies \cite{franceschelli2024creative, meister-etal-2023-locally}. In this work, we instead explore directly optimizing language models for creativity using human preferences extracted from responses to creativity assessments.

\vspace{-0.5mm}

\subsection{Preference Learning}
Aligning LLMs to human preferences has proven effective in developing models that are helpful and useful to users, leading to the emergence of numerous preference learning methods \cite{gao2024unifiedviewpreferencelearning, ouyang2022training, rafailov2023direct}. However, prior work has highlighted a lack of diversity in LLM outputs \cite{anderson2024homogenization,lanchantin2025diverse,wenger2025we,padmakumar2023does}, with alignment often cited as a contributing factor \cite{west2025basemodelsbeataligned}. In response, recent research has explored modifications to existing preference modeling techniques aimed at mitigating this reduction in diversity. One notable approach, Diverse Preference Optimization, proposes enhancing preference data creation by selecting preference pairs based on a diversity metric \cite{lanchantin2025diverse}. Another recent method introduces a modification to the optimization objective itself to incorporate a diversity signal \cite{chung2025modifying}. Both strategies have demonstrated effectiveness in promoting output diversity with minimal impact on output quality. However, as previously noted, diversity represents only one facet of creativity; true creativity also requires the capacity for novelty and surprise. In this work, we present a modular preference alignment framework for creativity that enables direct optimization across multiple dimensions of creative expression.

%% file: 03_cdpo.tex
According to its three-criterion definition, creativity involves the generation of novel, high-quality, and surprising ideas \cite{simonton2012taking, boden2004creative, runco2012standard}. Moreover, creative outputs tend to be highly diverse across individuals \cite{anderson2024homogenization}. Therefore, to promote overall creativity in LLM outputs, we propose to inject unsupervised metrics related to each dimension of creativity into the loss functions of standard preference optimization methods. We use direct preference optimization (DPO) \cite{rafailov2023direct} to illustrate our modifications to the loss function. Recall that in the standard formulation of DPO, a policy model ($p_\theta$) is directly optimized on a dataset of $(x, y^w, y^l)$ where $x$, $y^w$ and $y^l$ refer to the model input (i.e. prompt), preferred (i.e. chosen) model response and dispreferred (i.e. rejected) model response, respectively. Using the ratio between the policy model’s likelihood and that of the reference SFT model ($p_{SFT}$) as an implicit reward, the training objective of DPO is defined as follows \cite{rafailov2023direct}:

{\footnotesize
\begin{equation}
\begin{aligned}
l_{DPO} = \Bigl[
  \log \sigma \bigl(
    \beta \log \tfrac{p_\theta(y^w \mid x)}{p_{\mathrm{SFT}}(y^w \mid x)}
    - \beta \log \tfrac{p_\theta(y^l \mid x)}{p_{\mathrm{SFT}}(y^l \mid x)}
  \bigr)
\Bigr] \\
\mathcal{L_{DPO}} = - \mathbb{E}_{(x, y^w, y^l) \in D} \quad l_{DPO} \qquad \qquad \qquad \qquad
\end{aligned}
\end{equation}
}

A challenge with standard preference optimization methods is that they may significantly reduce the diversity of the responses LLMs generate, as the loss function encourages models to generate preferred responses even if they are not very creative \cite{west2025basemodelsbeataligned, padmakumar2023does, anderson2024homogenization, kirk2023understanding, xu2024echoes, o2024attributing, zhang2024forcing, wenger2025we}. Existing approaches to address this in the preference optimization objective have centered around curating preference data based on various diversity metrics \cite{lanchantin2025diverse} or incorporating extra regularization terms that encourage diverse generations while balancing quality \cite{chung2025modifying}. For example, the recently proposed Diversified DPO (DDPO) method adds a scalar diversity term $\delta^w$ (i.e. diversity score of the preferred response) into the DPO loss \cite{chung2025modifying}:

{\footnotesize
\begin{equation}
\begin{aligned}
\mathcal{L_{DDPO}} = -\mathbb{E}_{(x, y^w, y^l) \in D} \quad \delta^{w} l_{DPO}
\end{aligned}
\end{equation}
}

While diversity is important for creativity, research in psychology has long established that truly creative responses also require novelty, surprise, and quality \cite{boden2004creative, barron1955disposition, simonton2018defining}. Therefore, we propose incorporating metrics for each of these, alongside diversity, into the preference loss in a \textit{modular} structure, enabling the construction of different creativity models by combining these dimensions as needed.

{\footnotesize
\begin{equation}
\begin{aligned}
\mathcal{L_{CDPO}} = -\mathbb{E}_{(x, y^w, y^l) \in D} \Bigl[\qquad \qquad \qquad \\
 \quad (\lambda_d \delta^w + \lambda_n \nu^w + \lambda_s \xi^w + \lambda_q \gamma^w) l_{DPO} \Bigr]
\end{aligned}
\end{equation}
}

In our proposed creative DPO loss, $\delta^{w}$, $\nu^{w}$, $\xi^{w}$ and $\gamma^w$ correspond to diversity, novelty, surprise and quality scores of the preferred response respectively and $\lambda_{d}$, $\lambda_{n}$, $\lambda_{s}$ and $\lambda_{q}$ are hyperparameters that control the effect of each score (we call them injection weights). In particular, when $\lambda_{d} = 1$, $\lambda_{n} = 0$, $\lambda_{s} = 0$ and $\lambda_{q} = 0$, we recover the DDPO loss. While there are multiple approaches for operationalizing $\delta^{w}$, $\nu^{w}$, $\xi^{w}$ and $\gamma^{w}$, we propose to use the following metrics for each:

\subsection{Diversity}
\ME{Diversity is defined as the pairwise differences between artifacts, and semantic distance is typically used to measure the difference \cite{chung2025modifying, ismayilzada2024evaluating, padmakumar2023does}.} We use an inverse homogenization metric from \citet{padmakumar2023does} similar to \citet{chung2025modifying}. Specifically, given a prompt $x$ and a set of (preferred) responses for $x$ denoted as $Y_x$, we compute the diversity score of any particular preferred response as the average pairwise semantic distance to all the other preferred responses in $Y_x$:

{\footnotesize
\begin{equation}
    \delta^w = \frac{1}{|Y_x| - 1} \sum_{y_i \in Y_x \setminus y^w} \textit{semdis}(y^w, y_i)
\end{equation}
}

We use $1-cos\_sim(\cdot,  \cdot)$ as a semantic distance function (i.e., $semdis(\cdot, \cdot)$).

\subsection{Novelty}
\ME{Novelty is typically defined as the measure of how different an artifact is from other known artifacts in its class \cite{maher2010evaluating}, and semantic distance-based metrics have been established as a good proxy in creativity research \cite{Johnson2022DivergentSI, beaty2021automating, dunbar2009creativity, harbinson2014automated, Karampiperis2014}.} We use a novelty metric similar to \citet{Karampiperis2014} where the novelty of a text is defined as the absolute difference between the average pairwise semantic distances of words in the text and those of a reference corpus of texts. In particular, we define the set of preferred responses to a prompt $x$ as the reference corpus ($Y_x$) and define the novelty of a preferred response as follows:

{\footnotesize
\begin{equation}
    \nu^{w} = |DSI(y^{w}) - DSI(Y_x)|
\end{equation}
}

{\footnotesize
\begin{equation}
DSI(T) = \frac{\sum_{i,j=1}^{|T|} \textit{semdis}(T_i, T_j),\ i \ne j}{|T|}
\end{equation}
}

Here $T$ refers to a piece of text, $T_i$ to the word $i$ in the set of unique words in $T$ denoted as $|T|$, and $DSI(\cdot)$ is \textit{divergent semantic integration}, the average pairwise semantic distances of words in a text \cite{Johnson2022DivergentSI}. 


\subsection{Surprise}
\ME{Surprise or unexpectedness has many definitions in the cognitive science literature \cite{MODIRSHANECHI2022102712}, but it is generally characterized by deviation from the expected.} We use Shannon surprise -- the negative log-likelihood of the text --- which has been widely used as a measure of surprise in prior work \cite{Bunescu2022DistributionBasedMO, MODIRSHANECHI2022102712, Kuznetsova2013UnderstandingAQ}. More specifically, given a prompt $x$, we define the surprise of a particular response as the exponentiated negative log-likelihood of the response (i.e. perplexity) conditioned on the prompt $x$ and under some reference model $S$ as follows:

{\small
\begin{equation}
    \xi^{w} = 2^{-logP_S(y^{w}|x)}
\end{equation}
}

\subsection{Quality}
Although a general quality scoring method is hard to define \ME{as it is highly domain-dependent}, reward models that are trained to output a high score to preferred answers \ME{are increasingly being used to assess the overall quality of language model outputs and align them to human preference} \cite{zhang2025noveltybench, lambert2024rewardbench}. In particular, we define the quality of a preferred response given a prompt $x$ as the score assigned by some reward model $R$: $\gamma^w = R(y^w | x)$.

%% file: 04_MUSE.tex
To compile \MUSE{}, we solicited data from the global creativity research community, specifically targeting researchers studying human creativity to obtain data from tasks known to be valid creativity measures. We specifically targeted datasets which contained complete metadata, including information about the task, language, and items that participants responded to. We gathered additional data by performing a manual search of the Open Science Framework database\footnote{https://osf.io/}, and only retained data from peer-reviewed articles. In total, 43\% of the data in \MUSE{} has never been publicly released, making it unlikely that LLMs have seen the item-response combinations for the majority of our tasks.


Every response in \MUSE{} was rated for creativity by at least two raters, and in some cases up to 75 employing a missing-raters design \cite{forthmann2025planning}. While it is common practice to measure creativity using multiple independent raters, individual raters may deliver unhelpful or noisy ratings if they did not understand the task instructions, had a different understanding of the rating criteria, or for other reasons \cite{forthmann2017missing}. To account for this, we followed best practices for subjective scoring tasks by employing Judge Response Theory \cite{myszkowski2019judge} to check for raters whose ratings were uninformative in an information-theoretic sense. We fit JRT models to each task within \MUSE{}, which gave us an information function for each rater across tasks. We then input the results from the JRT into a genetic algorithm \cite{schroeders2016meta} which identified a subset of raters per dataset that maximized the per dataset rater information function.\footnote{While ensuring that the algorithm kept at least two raters per dataset.} This process dropped uninformative raters from each dataset, enhancing the quality of the final creativity ratings. The individual rater's scores were aggregated via factor scores, as is best practice in creativity assessment \cite{silvia2011subjective}, and we rescaled the factor-transformed creativity scores into the integer range 10-50 as is done for prior work in automated creativity assessment \cite{organisciak2023beyond}. 
Full details about the dataset construction are in Appendix \ref{sec:oracl_dataset}.

%% file: 05_experiments.tex
\begin{figure*}[t]
    \centering   
    \includegraphics[width=\linewidth,trim={0cm 0cm 0cm 0cm}]{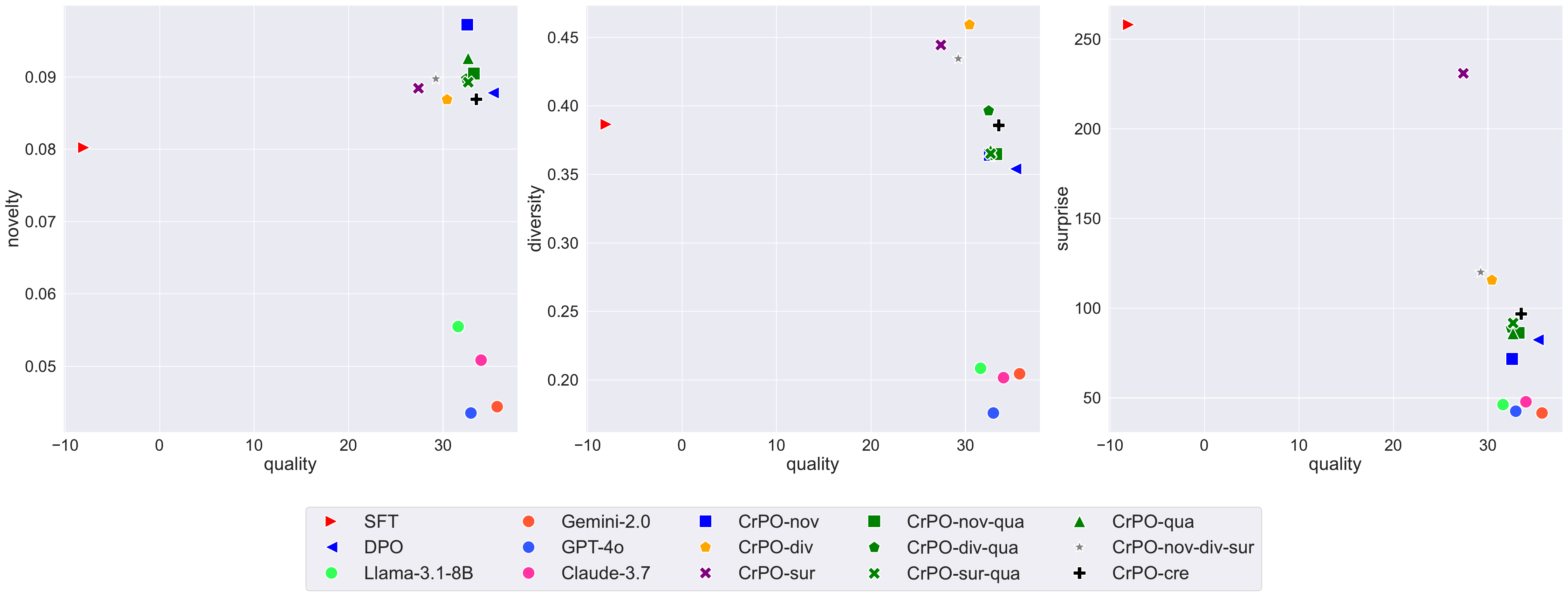}
    \caption{Results on held-out evaluation suite from \MUSE{} across all baselines and our models using \texttt{Llama-3.1-8B-Instruct} as a base model. \texttt{nov}, \texttt{div}, \texttt{sur}, \texttt{qua}, \texttt{cre} denote novelty, diversity, surprise, quality, and creativity, respectively. Results are averaged across tasks. \texttt{Mistral-7B-Instruct-v0.3} results can be found in Appendix Figure \ref{fig:main-mistral-results}.}
    \label{fig:main-results}
\end{figure*}

\subsection{SFT and Preference Datasets}
While our \MUSE{} dataset contains samples for multiple languages, we focus on showing the effectiveness of \CRPO{} on the English subset in this work and leave experiments using the full dataset as future work. From the base English \MUSE{} dataset, we generate a preference dataset by creating tuples of preferred and rejected responses to the same prompt, treating the response that received the higher creativity score as the preferred one. Past work has shown that data quality is one of the main factors behind preference model performance \cite{liu2024skywork, deng2025less, wang2024reward}. Therefore, we curate a high-quality SFT dataset of $5,285$ samples (\MUSE{}-\textsc{SFT}) and preference dataset of $42,058$ samples (\MUSE{}-\textsc{Pref}) from the base \MUSE{} which we detail in Appendix \ref{sec:app-datasets}.

\subsection{Training}
\paragraph{Models}
As our base models, we use \texttt{Llama-3.1-8B-Instruct} \cite{llama3modelcard} and \texttt{Mistral-7B-Instruct-v0.3} \cite{jiang2023mistral7b} and implement \CRPO{} as described in Section \ref{sec:cdpo}. We first train our models using supervised fine-tuning (SFT model) for a single epoch on \MUSE{}-\textsc{SFT}, and then apply preference optimization on the SFT model using \CRPO{} and \MUSE{}-\textsc{Pref} dataset. We train all models using parameter-efficient tuning with LoRA using a rank of $128$ and an alpha of $256$ \cite{hu2022lora}. Additional details on the training setup can be found in Appendix \ref{sec:app-training-details}.

\begin{figure*}[t]
    \centering   
    \includegraphics[width=\linewidth,trim={0cm 0cm 0cm 0cm}]{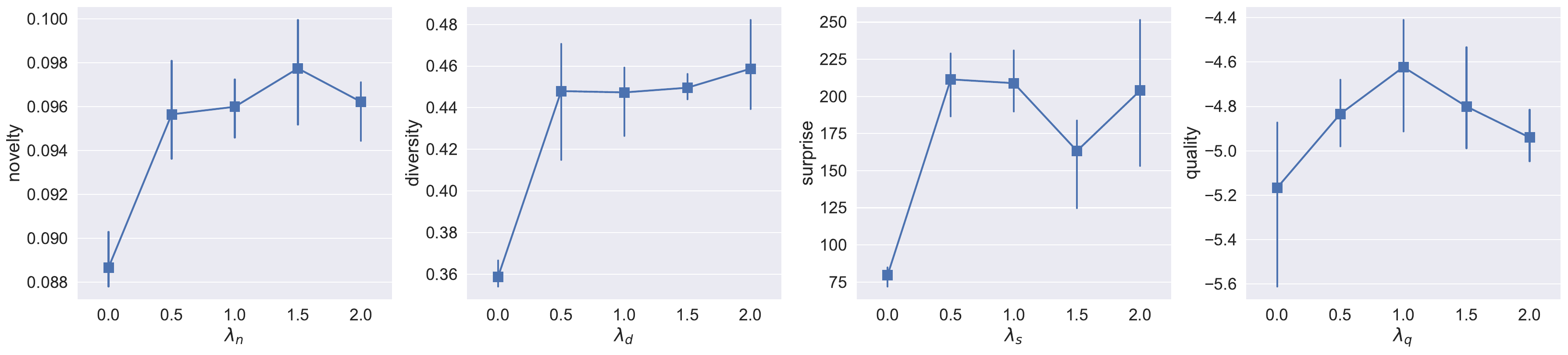}
    \caption{Effect of injection weights for each dimension. Results are averaged across three seed runs.}
    \label{fig:lambda-results}
\end{figure*}

\paragraph{Creativity Injection}
 We compute creativity metric scores for each preferred response and inject them into the DPO objective function as described in Section \ref{sec:cdpo}. Since each metric is on a different scale and we would like to combine the effects of different injections, we normalize each score to a range of $[0,1]$ before injection. We vary the injection weights $\lambda_d, \lambda_n, \lambda_s, \lambda_q$ accordingly\footnote{For example, to train a novelty model, we set $\lambda_n=1$ and others to 0 whereas for novelty and quality model we set $\lambda_n=1$ and $\lambda_q=1$.} to train different suites of creative models. As novelty and diversity measures require a reference set to compute against, we adopt a prompt-level granularity and consider the set of responses for a given prompt as the reference corpus similar to prior work \cite{chung2025modifying}. We use the \texttt{jina-embeddings-v3} model \cite{sturua2024jinaembeddingsv3multilingualembeddingstask} to compute text embeddings for all metrics that rely on semantic distance. 
 For surprise, we use instruction-tuned \texttt{Gemma-2-27B} \cite{gemmateam2024gemma2improvingopen} as our reference surprise model $S$. While our creativity preference dataset is already high-quality, we also experiment with injecting external quality signals to study its interaction with other creativity dimensions. Hence, for the quality measure, we employ an existing reward model \texttt{Skywork-Reward-Gemma-27B-v0.2} \cite{liu2024skywork} that is one of the top-performing models on RewardBench \cite{lambert2024rewardbench} as our reference reward model $R$.

\subsection{Evaluation}
\paragraph{Tasks and Metrics}
We evaluate all models across several dimensions of creativity on held-out prompts of various tasks and two held-out tasks. More specifically, we use $5$ held-out prompts from \textit{Real-Life Creative Problem Solving, Alternate Uses of Objects, Design Solutions, Hypothesis Generation,} and \textit{Metaphors} tasks, and $9$ prompts from two held-out tasks of \textit{Poems} and \textit{Sentence Completion}. For each prompt, we generate $16$ responses from each model by varying the \texttt{$temperature$}, \texttt{$top_p$}, and \texttt{$top_k$} decoding parameters. Our final held-out evaluation suite contains $224$ samples. We evaluate the responses on the dimensions of novelty, diversity, and surprise using the metrics described in Section \ref{sec:cdpo}. Additionally, to study the tradeoff between creativity and quality, we train a reward model on our preference dataset using instruction tuned \texttt{Gemma-2-9b} \cite{gemmateam2024gemma2improvingopen} and use it to score the overall quality of model generations. More details about the evaluation setup can be found in Appendix \ref{sec:app-eval-details}.

\paragraph{Baselines} 
As baselines \footnote{\ME{We note that DDPO \cite{chung2025modifying} is a special case of our method, and the model trained with it is equivalent to our \CRPO{}\texttt{-div} variant. For this reason, we do not treat it as a separate baseline. Moreover, since DDPO has already been shown to outperform the Diverse Preference Optimization method \cite{lanchantin2025diverse}, a direct comparison is unnecessary by transitivity.}}, we use the base models \texttt{Llama-3.1-8B-Instruct} \cite{llama3modelcard} and \texttt{Mistral-7B-Instruct-v0.3} \cite{jiang2023mistral7b}, SFT models which are the base models supervised fine-tuned on \MUSE{}-\textsc{SFT}, a vanilla DPO model trained on top of the SFT model using the \MUSE{}-\textsc{Pref} dataset without any creativity injections and three closed-source instruction-tuned LLMs, namely \texttt{GPT-4o} \cite{openai2024gpt4ocard}, \texttt{Claude-3.7-Sonnet} \cite{claude37}, and \texttt{Gemini-2.0-Flash} \cite{gemini20}. \ME{We also consider additional baselines such as \textit{"Brainstorm, then select"}, a creative prompting approach that has shown improvement on creativity scores in the Alternative Uses of Objects Task (AUT) \cite{SummersStay2023BrainstormTS} and \textit{min-p sampling}, a decoding strategy that directly promotes output creativity \cite{nguyen2024turning}. Evaluation details for these baselines can be found in Appendix \ref{sec:app-additional-baselines}.}

\paragraph{\CRPO{} Models}
We train several \CRPO{} models corresponding to the different dimensions of creativity. More specifically, for each dimension, we train a model that is injected with a signal for the given dimension and another model that is injected with a signal for both the given dimension (e.g. \CRPO{}\texttt{-nov}) and the quality dimension (e.g. \CRPO{}\texttt{-nov-qua}). We train the latter models to understand the tradeoff between other dimensions of creativity and the quality that has been reported in previous research \cite{zhang2025noveltybench, lanchantin2025diverse, chung2025modifying}. Additionally, we train two creative models that inject all dimensions of creativity (denoted as \CRPO{}\texttt{-cre}) and all except quality (denoted as \CRPO{}\texttt{-nov-div-sur}). In all these experiments, $\lambda$ injection weights are set to $1$ for simplicity. We perform a more detailed analysis of these hyperparameters in Section \ref{sec:effect-lambda}.

%% file: 06_results.tex
Figure \ref{fig:main-results} summarizes performance on our held-out evaluation suite across creativity dimensions for all baselines and \CRPO{} models using the \texttt{Llama-3.1-8B-Instruct} as a base. Results for \texttt{Mistral-7B-Instruct-v0.3} can be found in Appendix Figure \ref{fig:main-mistral-results} and follows the same trends. First, we observe a clear separation between existing instruction-tuned LLMs and our models: while the former cluster around high quality but low novelty, diversity, and surprise, our models achieve high scores across all four dimensions. Second, for each creativity dimension, the model trained with that specific injection outperforms others on the same metric, confirming the effectiveness of targeted optimization, without a considerable drop in quality.

Models that combine a creativity signal with an external quality signal (\CRPO{}\texttt{-\{nov,div,sur\}-qua}) improve in quality but show reduced performance on the targeted dimension, illustrating a trade-off. The same pattern holds when comparing the \CRPO{}\texttt{-nov-div-sur} model to the full \CRPO{}\texttt{-cre} model, further highlighting the balance between quality and other facets of creativity. Interestingly, the vanilla DPO model, without any creativity injections, already outperforms existing LLM baselines, demonstrating the strength of our preference dataset. Still, most of our creativity-optimized models significantly surpass DPO across all dimensions. Finally, the SFT model performs worst in quality and shows only comparable performance on other dimensions, reinforcing prior findings \cite{chung2025modifying} about the limited generalizability of supervised fine-tuning in creative tasks, where no single ``correct'' answer exists.

\ME{We also compare our approach to creative prompting (Brainstorm \& Select) and decoding (\textit{min-p} sampling) strategies which we detail in Appendix \ref{sec:app-additional-baselines} and results are reported in Appendix Tables \ref{tab:res-creative-prompt} and \ref{tab:res-creative-decoding}. We can see that while the creative prompting and decoding strategies improve baseline performance, our models with the standard prompting and decoding approaches still beat them across all dimensions with minimal drop in quality.}

Overall, \textbf{our results show that CrPO enhances multiple aspects of creativity with minimal impact on quality}, offering a flexible and effective framework for creativity alignment in LLMs.

\subsection{Effect of Injection Weights}
\label{sec:effect-lambda}
While we set all injection weights to 1 for simplicity in our main evaluations, we also study the effect of the different injection values on the performance of models across dimensions. In particular, we vary the injection weights from $0$ to $2.0$ with an increment of $0.5$ for all dimensions and report the averaged results across three seed runs in Figure \ref{fig:lambda-results}. We observe that across most dimensions, an injection weight of $0.5$ yields the greatest performance gains, with further increases resulting in diminishing returns or slight performance degradation. In terms of quality, the injection weight of $1.0$ results in the highest performance. \ME{To gain further insights into the inherent interactions between creativity dimensions, for each dimension, we also report the performance of a model optimized for a given dimension on all the other dimensions and across increasing values of injection weights. Results for this analysis can be found in Appendix Figures \ref{fig:lambda-results-novelty-int}, \ref{fig:lambda-results-diversity-int}, \ref{fig:lambda-results-surprise-int}. Results show that in general, the interaction between different creativity dimensions is complex with some discernible patterns. One clear trend is that the increase in novelty, diversity, and surprise injections results in a quality drop, though not significant. Interestingly, we observe that while an increase in novelty injection follows with overall increase in surprise performance, the opposite does not seem to be true and in fact, more surprise injection seems to cause a drop in novelty. On the other hand, novelty and diversity dimensions seem to have a mutual positive correlation. Finally, we observe no clear relationship between the dimensions of surprise and diversity.} In general, we suggest tuning the injection weights depending on the training dataset, underlying task, and the base model for the best performance.

\subsection{Human Evaluation}
\label{sec:human-eval}
In addition to automated metrics, we conduct a human evaluation to assess the real-world effectiveness of our approach. Due to the high cost of human studies, we focus on the overall creativity dimension using a single task (Sentence Completion), 4 prompts, 4 baselines (SFT, DPO, \texttt{Llama-3.1-8B-Instruct}, and \texttt{GPT-4o}), and 5 \CRPO{} variants (\texttt{{nov, div, sur, nov-div-sur, cre}}). In a blind pairwise setup, participants compared responses from a baseline and a \CRPO{} model for creativity, unaware that the texts were AI-generated. A total of 320 comparisons were collected with balanced sampling across models. Additional details are in Appendix \ref{sec:app-human-studies}.

Figure \ref{fig:human-eval-cre-results} presents the win rates. The \CRPO{}-\texttt{nov-div-sur} model consistently outperforms all baselines, particularly \texttt{Llama-3.1-8B-Instruct}, by a wide margin. In contrast, the full \CRPO{}-\texttt{cre} model lags slightly, reflecting the creativity–quality tradeoff seen in automated evaluations. Notably, \CRPO{} models achieve especially strong gains over SFT, reinforcing previous findings.

\begin{figure}[t]
    \centering   
    \includegraphics[width=\linewidth,trim={0cm 0cm 0cm 0cm}]{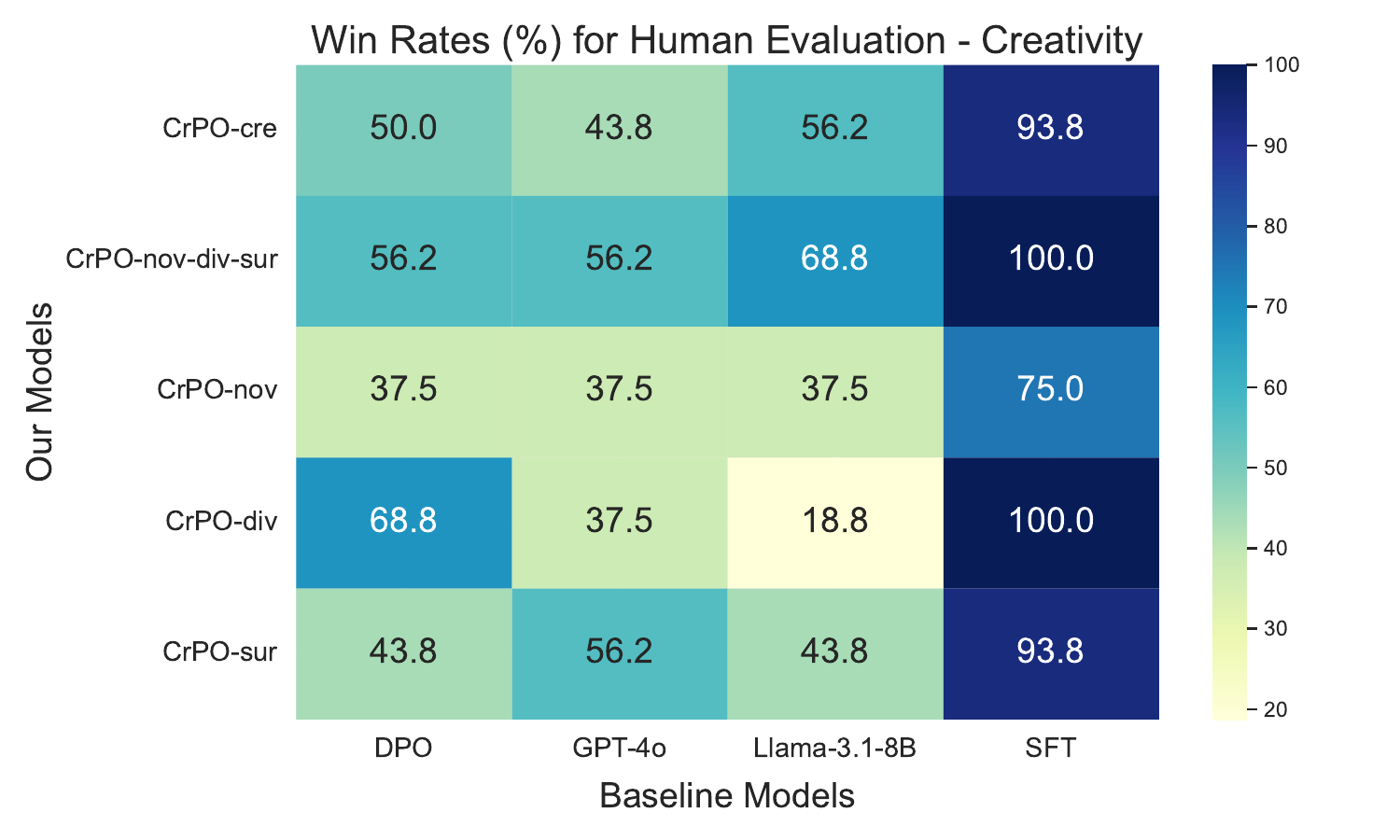}
    \caption{Human evaluation results measured by win rates. Participants were asked to make a pairwise comparison between our models and baselines with respect to the overall creativity.}
    \label{fig:human-eval-cre-results}
\end{figure}

\begin{figure}[h]
    \centering   
    \includegraphics[width=\linewidth,trim={0cm 0cm 0cm 0cm}]{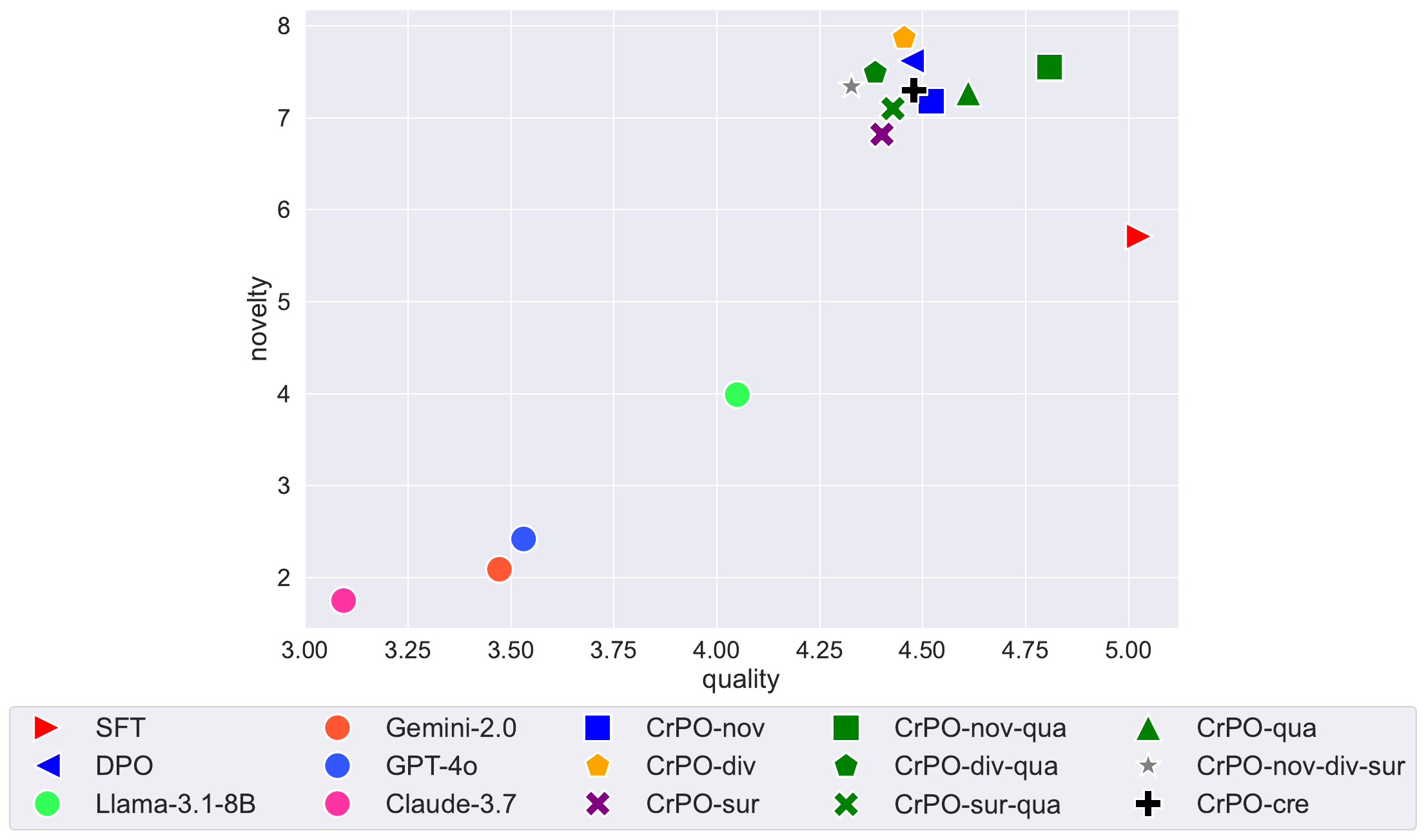}
    \caption{Evaluation results on \textsc{NoveltyBench}, using the novelty and quality metrics defined in \citet{zhang2025noveltybench}.}
    \label{fig:nbench-results}
\end{figure}

\subsection{\textsc{NoveltyBench} Evaluation}
\label{sec:nbench-eval}
While we demonstrate the effectiveness of our approach on the \MUSE{} held-out set using automated metrics, we also evaluate generalization on external benchmarks using the recently introduced \textsc{NoveltyBench} \cite{zhang2025noveltybench}. This benchmark includes tasks spanning randomness, factual knowledge, creative writing, and subjectivity. Following the recommended evaluation setup, we benchmark all baselines and \CRPO{} variants on a curated 100-prompt subset, using the benchmark’s novelty and quality metrics. Full details are in Appendix \ref{sec:app-nbench-details}.

Figure \ref{fig:nbench-results} shows novelty vs. quality scores across all models and tasks. As in our internal evaluation, we observe a clear separation: existing LLM baselines cluster around lower novelty and variable quality, while our models consistently achieve high scores on both dimensions. Notably, although our models outperform SFT on novelty, the SFT model surprisingly achieves higher quality—beating both baselines by a large margin and our models by a smaller one. This aligns with findings from \textsc{NoveltyBench} \cite{zhang2025noveltybench}, where smaller models like \texttt{Gemma-2-2B-it} and \texttt{Llama-3.1-8B-Instruct} often surpass larger ones in quality.

Overall, our models set a new state-of-the-art on the \textsc{NoveltyBench} leaderboard in terms of novelty. \footnote{https://novelty-bench.github.io/}

%% file: 07_discussion.tex
\ME{In this work, we propose a modular preference alignment framework that uses signals from different dimensions of creativity to promote overall output creativity improvement. We show that small LLMs trained using our framework on a psychologically-valid datasets of human creativity outperform strong baselines such as GPT-4o on both automated and human evaluations. Our formulation builds on well-established dimensions from creativity research—specifically novelty, surprise, diversity, and quality—which are frequently identified as core components in both cognitive psychology \cite{MODIRSHANECHI2022102712, grace2016surprise, barron1955disposition} and computational creativity literature \cite{runco2012standard, simonton2012taking, boden2004creative}. We also employ creativity metrics that provide measurable signals aligning with key cognitive theories and enable practical optimization within language models. These signals have also been shown to provide a good feedback mechanism and increase human creativity in several tasks \cite{de2023automated, de2025artificial}. Similarly, our work shows that these signals can be used as an implicit feedback mechanism in the learning process to improve language model creativity.
Additionally, our framework is well-suited for downstream creative tasks where certain dimensions of creativity matter more than others. In creative writing, novelty and diversity ensure fresh, engaging content, while surprise drives plot twists. In advertising, novelty and surprise capture audience attention in crowded markets. For brainstorming, diversity is key to exploring a wide idea space. By enabling models to be trained with customizable mixes of creativity dimensions, our framework adapts to the needs of each domain.}

%% file: 08_conclusion.tex
We introduce \CRPO{}, a flexible methodology for enhancing the creativity of LLMs. Leveraging a novel large-scale human preference dataset focused on creativity, we show that models aligned with \CRPO{} produce generations that are not only novel, diverse, and surprising, but also high in quality —-- on both our held-out evaluation suite and the external \textsc{NoveltyBench} dataset. Human evaluations further confirm that raters consistently judge our model’s outputs to be more creative than those of several strong baselines, highlighting the potential of our approach to boost LLM creativity. While our experiments focus on smaller models such as \texttt{Llama-3.1-8B} and an English-only dataset, future work could explore the scalability of \CRPO{} to larger models, multilingual settings and other preference optimization methods.

%% file: 09_limitations.tex
\section*{Limitations}

Due to constraints on both computational resources and budget for human studies, we were unable to evaluate \CRPO{} on any languages other than English. Multilingual creativity assessment using generative AI remains a challenging problem and an active area of research \cite{haase2025sdatmultilingualgenaidrivenframework}. While we believe our data represents a valuable resource for the community, future work will need to test our methods in multilingual settings to ensure multilingual generalization. These compute constraints also prevented us from evaluating \CRPO{} on larger open-weight models, making scaling trends difficult to predict. We retained only samples with full agreement for the creativity score when training our models. While this aligns with best practices for creativity measurement in psychology \cite{cseh2019scattered}, it may also mask genuine sources of rater disagreement that should be modeled. Moreover, we acknowledge that, much like other datasets used to align LLMs, the preferences represented by our annotator population likely do not reflect the full range of human preferences, which could bias our models' generations \cite{yeh2024reliable}. We believe that the large-scale and multilingual nature of our collected data likely makes it one of the most representative creativity datasets currently available, but stress that future work should consider issues of bias and fairness more carefully for LLM creativity assessment. Finally, we also acknowledge that creativity is a complex and subjective construct, and the metrics we used may not capture the full richness of human creativity judgments.

\section*{Ethical Considerations}
We emphasize that our models should not be used for safety-critical applications, as the relationship between creativity and alignment with other values remains underexplored. Notably, our dataset contains responses to tests of malevolent creativity that are by definition unsafe for models to generate. We also observed qualitatively that \CRPO{} models were more likely to generate unsafe or toxic responses even to prompts that did not explicitly request such behaviors. We believe that our data is valuable for red-teaming evaluations on tasks requiring creativity, and that aligning models on these malevolent responses could be beneficial for understanding how malicious actors might use creativity-enhanced models to execute unsafe goals. However, we also acknowledge the ethical concerns that the release of our models and datasets would raise, and believe that restricting access to only those which have signed a license agreement is the best approach for balancing safety with continued scientific advancement. While we believe our results demonstrate how aligning LLMs with carefully designed human creativity datasets can significantly improve the novelty and diversity of their generations, it remains unclear how to both optimize for creativity while preserving guardrails that prevent unsafe behavior.

We also acknowledge the broader debates around the valid use of AI in social-behavioral research \cite{sun2025sociodemographic} and concerns surrounding AI automation of industries requiring creativity \cite{Wilkinson_2023} in which our work is situated. While the over-reliance on AI for creative tasks to the detriment of human welfare is a legitimate concern, AI has also been acknowledged for its potential to enhance human creativity above and beyond what might be possible otherwise \cite{deartificial, loi2020blindspot, loi2020societal}. Creativity is a vital skill for future knowledge workers to master \cite{World_Economic_Forum_2025a}, and we believe that enhancing the creativity of AI is an important prerequisite for developing AI systems capable of training humans to be more creative.



%% file: acknowledgements.tex
Mete and Lonneke gratefully acknowledge the support of the Swiss National Science Foundation (grant 205121\_207437: C - LING) and Fondazione Aldo e Cele Daccò. Antoine Bosselut gratefully acknowledges the support of the Swiss National Science Foundation (No. 215390), Innosuisse (PFFS-21-29), the EPFL Center for Imaging, Sony Group Corporation, and a Meta LLM Evaluation Research Grant. Roger E. Beaty is supported by grants from the US National Science Foundation [DRL-1920653; DRL-240078; DUE-2155070].

%% file: 10_appendix.tex
\section{\MUSE{} Dataset}
\label{sec:oracl_dataset}
We compiled data by means of crowdsourcing and data mining of the open-source data sharing platform OSF. We crowdsourced from the global creativity research community by means of direct requests and posts on academic listservs. In our call for data-sharing, we requested data relating to any creativity responses that were provided by human participants and scored for creativity by human raters. We specifically requested that the datasets include scores from each rater, rather than composite creativity scores, to determine rating data quality for each submission. As part of our inclusion criteria, we further requested that researchers provide information relating to: (a) the creativity task, (b) the item associated with each response, (c) the construct that was rated, and (d) the language of the task. We further asked researchers to provide a statement on whether they agreed to making their data open-source. In terms of data mining through the OSF platform, we first searched through a series of relevant keywords (e.g., “creativity task”, “originality score”). We only retained sub-datasets from credible sources, which were associated with a citable peer-reviewed article, and which included all the required data relating to our inclusion criteria.

After removing responses that didn’t meet our inclusion criteria, our dataset amounted to 321,572 human-rated and language-based creativity responses. The dataset was thus cleaned by standardizing the naming for each variable except for the responses. We then removed responses for having been rated by fewer than 2 human judges. Duplicate responses were also removed, by retaining a single exemplar for responses that appeared twice within a specific item and task. 

To enhance the reliability of human creativity ratings across the numerous datasets, we optimized the selection of raters by applying a meta-heuristic algorithm. Specifically, we applied a Genetic Algorithm \cite{schroeders2016meta}. The GA operates through iterative selection, crossover, and mutation processes, mirroring the principles of natural selection, and in our case to identify the optimal subsets of raters for each dataset. In each iteration, candidate solutions—that is, combinations of raters—were evaluated based on a predefined fitness function that prioritized the maximization of empirical reliability (rxx) within a graded response model (GRM) and hence in line to judge response theory. For sub-datasets involving decimal-based scales, individual ratings were rounded to the nearest integer value (rounding up if containing a decimal = .5) to meet the requirements of the GRM.

Rater subsets demonstrating superior reliability were selected, recombined, and modified through random perturbations to prevent premature convergence to suboptimal solutions. This approach ensured that the selected raters provided consistent and informative judgments while reducing noise introduced by inconsistent or uninformative ratings. By automating the selection process through GA, we opted for maximal comparability in the selection process across datasets. Previous research has demonstrated the utility of GA in psychometric optimization tasks, particularly in balancing brevity and measurement precision while maintaining construct validity. In the present study, GA facilitated a systematic and data-driven refinement of rater selection, arguably enhancing the overall quality of creativity ratings.

After dropping uninformative raters in each sub-dataset, we again removed any rows containing less than 2 ratings due to rater removal. Afterwards, we used the new rater subsets per dataset and computed factor scores for each given response that were used as creativity scores. We calculated factor scores via a GRM model, ran separately over each sub-dataset, to derive a single creativity score for each response. Finally, we applied min-max scaling on each sub-dataset to transform ratings into a range of 10 to 50, with intervals of 1. This step was applied to ensure that ratings would only constitute a single token in length, to lessen the burden of predicting multi-token labels by the LLMs.

We then withheld all responses in the Spanish language from our final dataset and assigned them to an out-of-distribution-language (OOD-l) set. Responses from the OOD-l set were not included in the training data of \MUSE{}, allowing us to test whether the model could generalize to creative responses in an unseen language. We selected Spanish as it would allow for a fair test of generalizability given: (1) Spanish tends to be a high-resource language within the pre-training of modern LLMs, (2) it is similar to other Latin-root languages in our training data (e.g., Italian), (3) responses in Spanish spanned multiple creativity tasks, and (4) the language spanned a limited number of responses in our total dataset. We further withheld all responses from two highly-naturalistic tasks, the Poem and Alternative Title Generation, and assigned these to an out-of-distribution task (OOD-t) set. We selected these tasks as they made up a limited portion of the total dataset and would provide a test of \MUSE{}’s performance on unseen naturalistic creativity tasks.

We then randomly selected items within each task and assigned them to an out-of-distribution item (OOD-i) set. We identified candidate items that corresponded to 5\% or less of the responses within a task. Then, for tasks that contained 20 or more total items, we randomly assigned 2 of these items to our OOD-i set. For tasks that contained fewer than 20 total items, we instead randomly assigned 1 of these items to the OOD-i set. Finally, we split the remaining responses in our dataset into training, validation, and out-of-distribution responses (OOD-r) sets according to an 80/10/10 split. We grouped responses into unique combinations of sub-dataset, task, language, item, and rating label, then randomly assigned responses within each combination to each of the sets, ensuring an equal representation of responses associated with each of these variables within the training, validation, and OOD-r sets. Table \ref{tab:dataset_stas} contains the final dataset statistics for \MUSE{}. Tables \ref{tab:muse-task-desc} and \ref{tab:muse-task-desc2} contain the descriptions and data statistics for each task in \MUSE{}. Tables \ref{tab:muse-task-ex}, \ref{tab:muse-task-ex2}, \ref{tab:muse-task-ex3}, \ref{tab:muse-task-ex4}, and \ref{tab:muse-task-ex5} list some example prompts and low-rated and high-rated responses for each task from \MUSE{}.

\begin{figure*}[t]
    \centering   
    \includegraphics[width=\linewidth,trim={0cm 0cm 0cm 0cm}]{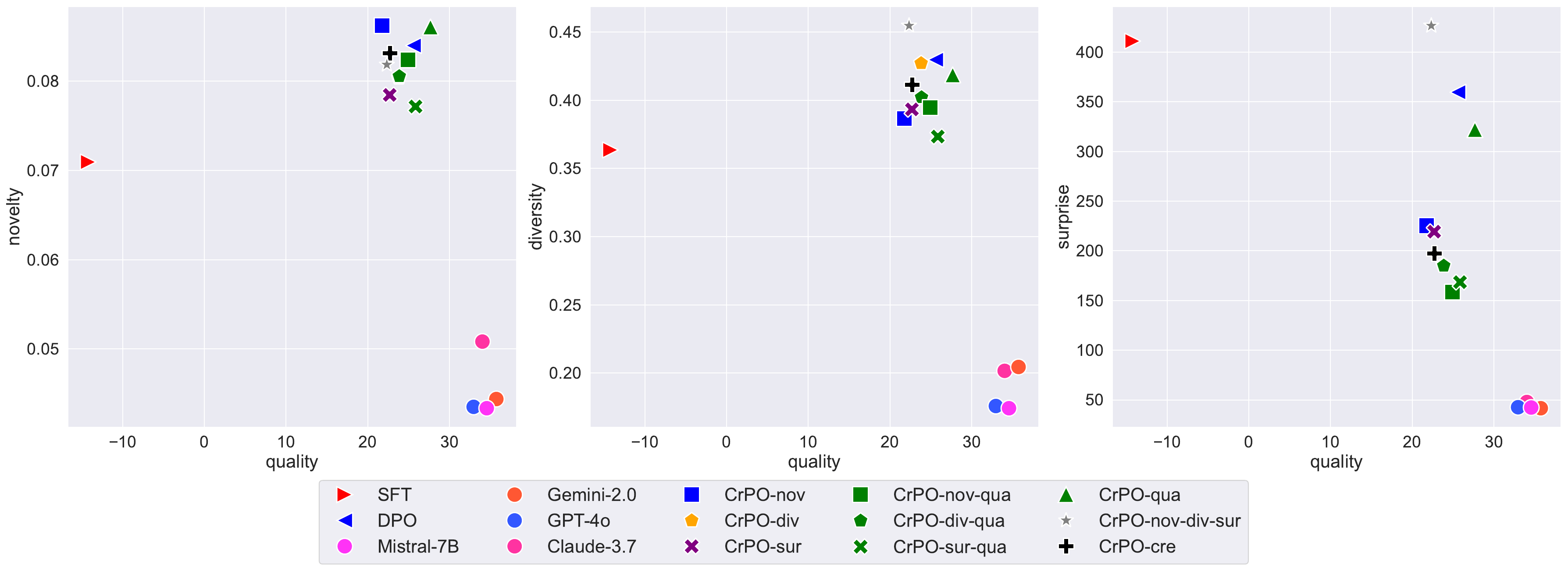}
    \caption{Results on held-out evaluation suite from \MUSE{} across all baselines and our models using \texttt{Mistral-7B-Instruct-v0.3} as a base model. \texttt{nov}, \texttt{div}, \texttt{sur}, \texttt{qua}, \texttt{cre} denote novelty, diversity, surprise, quality, and creativity, respectively. Results are averaged across tasks.}
    \label{fig:main-mistral-results}
\end{figure*}

\begin{figure}[h]
    \centering
    \begin{mybox}{Human Evaluation Instructions}
        In this study, you will be presented with two responses to a creative task. 
        Your job is to select the response that you believe is the most creative. 
        Please base your judgment only on the creativity of the ideas—not on how long or detailed the response is. A shorter response can be more creative than a longer one, and vice versa. Focus on how original, unique, and innovative the idea feels to you. 
        There are no right or wrong answers—we’re interested in your opinion.
    \end{mybox}
    \caption{Rater instructions for the human evaluation.}
    \label{fig:creativity-human-instructions}
\end{figure}

\begin{table*}[h]
    \centering
    \begin{tabular}{c c c c c c c c}
    \hline
         &  \textbf{Total} & \textbf{Train} & \textbf{Dev} & \textbf{Test} & \textbf{OOD-i} & \textbf{OOD-l} & \textbf{OOD-t} \\ \hline
      \# samples & 245,030 & 183,973 & 23,254 & 22,419 & 6,253 & 4,719 & 4,412 \\ 
      \# tasks & 25 & 23 & 23 & 23 & 9 & 3 & 2 \\
      \# languages & 11 & 10 & 10 & 10 & 5 & 1 & 3 \\
      \# prompts & 587 & 503 & 502 & 502 & 12 & 52 & 21 \\
      \hline
    \end{tabular}
    \caption{Detailed statistics for each split of \MUSE{}.}
    \label{tab:dataset_stas}
\end{table*}

\section{SFT and Preference Datasets}
\label{sec:app-datasets}
 Past work has shown that data quality is one of the main factors behind preference model performance \cite{liu2024skywork, deng2025less, wang2024reward}. In particular, the margin in the score (i.e. reward margin) between the preferred and rejected response may influence the performance of the model, since training pairs with smaller margins are likely to contain annotation noise and be more difficult to learn. We experiment with different reward margins and choose a margin of $5$ for the final experiments as it showed a balance between mitigating annotator noise and creating a dataset with nuanced preferences. Additionally, to ensure a high-quality preference dataset, first we filter the base \MUSE{} dataset and select only the samples that have a full agreement from all annotators. Then we filter out all samples that have a rating below $20$ and limit the number of pairings between samples to $10$. This results in a final preference training dataset of $42,058$ samples (\MUSE{}-\textsc{Pref}). We also create a high-quality instruction-tuning dataset from \MUSE{}-\textsc{Pref} by pairing the prompts with all preferred responses that have a rating above $30$ resulting in a dataset of $5,285$ samples (\MUSE{}-\textsc{SFT}). Tables \ref{tab:muse-pref-details} and \ref{tab:muse-sft-details} contain the statistics for these datasets.

\begin{table*}[h]
    \centering
    \begin{tabular}{c|c|c|c|c}
        \textbf{Model} & \textbf{novelty} & \textbf{diversity} & \textbf{surprise} & \textbf{quality} \\
         \hline
         \texttt{GPT-4o} (std. prompting) & $0.04$ & $0.18$ & $42.63$ & $\mathbf{32.94}$ \\
         \texttt{GPT-4o} (Brainstorm \& Select) & $0.05$ & $0.28$ & $14.26$ & $22.01$ \\
         \CRPO{}\texttt{-nov} & $\mathbf{0.10}$ & $0.43$ & $143.98$ & $27.46$ \\
         \CRPO{}\texttt{-div} & $\mathbf{0.10}$ & $\mathbf{0.49}$ & $204.48$ & $28.87$ \\
         \CRPO{}\texttt{-sur} & $\mathbf{0.10}$ & $\mathbf{0.49}$ & $\mathbf{590.65}$ & $17.91$ \\
         \CRPO{}\texttt{-nov-div-sur} & $\mathbf{0.10}$ & $0.47$ & $249.31$ & $27.07$ \\
         \CRPO{}\texttt{-cre} & $\mathbf{0.10}$ & $0.44$ & $203.17$ & $28.55$ \\
    \end{tabular}
    \caption{AUT evaluation results comparing a creative prompting strategy such as 
    Brainstorm \& Select, to our approach with standard prompting. Best performance is \textbf{bolded} for each metric.}
    \label{tab:res-creative-prompt}
\end{table*}

\section{Training}
\label{sec:app-training-details}
We follow a training setup similar to \citet{chung2025modifying} and use \texttt{Llama-3.1-8B-Instruct} and \texttt{Mistral-7B-Instruct-v0.3} \cite{jiang2023mistral7b} as our base models. Using these models, we train an SFT, DPO and several \CRPO{} models. We train all models using parameter-efficient tuning with LoRA using a rank of $128$ and an alpha of $256$ \cite{hu2022lora}. All training was done using HuggingFace TRL library\footnote{https://huggingface.co/docs/trl/en/index} with Accelerate \cite{accelerate} and DeepSpeed ZeRO-2 \cite{rajbhandari2020zero} on \texttt{NVIDIA A100} GPUs with gradient checkpointing.

\begin{table}[h]
    \centering
    \begin{tabular}{p{2cm} p{2cm} p{2cm}}
    \hline
        \textbf{Task} & \textbf{\# prompts} & \textbf{\# samples} \\
        \hline
        \textit{Real-Life Creative Problem Solving} & 8 & 5,601 \\
        \hline
        \textit{Question Asking} & 5 & 314 \\
        \hline
        \textit{Malevolent Problems} & 21 & 424 \\
        \hline
        \textit{Metaphors} & 51 & 675 \\
        \hline
        \textit{Alternate Uses of Objects Task} & 11 & 4,388 \\
        \hline
        \textit{Design Solutions} & 10 & 1,366 \\
        \hline
        \textit{Essays} & 1 & 174 \\
        \hline
        \textit{Stories} & 7 & 1,498 \\
        \hline
        \textit{Consequences} & 5 & 10,865 \\
        \hline
        \textit{Experiment Design} & 7 & 5,640 \\
        \hline
        \textit{Hypothesis Generation} & 6 & 5,260 \\
        \hline
        \textit{Research Questions} & 5 & 5,832 \\
        \hline
        \textit{Associations} & 5 & 21 \\
        \hline
        \textbf{Total} & \textbf{142} & \textbf{42,058} \\
      \hline
    \end{tabular}
    \caption{\MUSE{}-\textsc{Pref} training dataset details.}
    \label{tab:muse-pref-details}
\end{table}

\begin{table}[h]
    \centering
    \begin{tabular}{p{2cm} p{2cm} p{2cm}}
    \hline
        \textbf{Task} & \textbf{\# prompts} & \textbf{\# samples} \\
        \hline
        \textit{Real-Life Creative Problem Solving} & 8 & 642 \\
        \hline
        \textit{Question Asking} & 6 & 58 \\
        \hline
        \textit{Malevolent Problems} & 22 & 82 \\
        \hline
        \textit{Metaphors} & 60 & 158 \\
        \hline
        \textit{Alternate Uses of Objects Task} & 11 & 855 \\
        \hline
        \textit{Design Solutions} & 12 & 150 \\
        \hline
        \textit{Essays} & 1 & 23 \\
        \hline
        \textit{Stories} & 7 & 256 \\
        \hline
        \textit{Instances of Common Concepts} & 4 & 10 \\
        \hline
        \textit{Consequences} & 5 & 1,315 \\
        \hline
        \textit{Experiment Design} & 7 & 573 \\
        \hline
        \textit{Hypothesis Generation} & 6 & 548 \\
        \hline
        \textit{Research Questions} & 5 & 587 \\
        \hline
        \textit{Associations} & 7 & 28 \\
        \hline
        \textbf{Total} & \textbf{161} & \textbf{5,285 }\\
      \hline
    \end{tabular}
    \caption{\MUSE{}-\textsc{SFT} training dataset details.}
    \label{tab:muse-sft-details}
\end{table}

SFT model is trained on the \MUSE{}-\textsc{SFT} dataset for a single epoch with a batch size of $2$ per GPU using a gradient accumulation size of $4$ and context size of $1024$. We use a cosine scheduler with a half-cycle warmup and maximum learning rate of $3e-5$. Final model achieves $85\%$ mean token accuracy on the validation set.

DPO and \CRPO{} models are trained using the SFT model as a base on our \MUSE{}-\textsc{Pref} dataset for a single epoch with a batch size of $8$ per GPU using a gradient accumulation size of $8$ and context size of $1024$. We use a linear scheduler with a learning rate of $5e-6$. All final models achieve over $82\%$ reward accuracy on the validation set.

\begin{figure*}[h]
    \centering   
    \includegraphics[width=\linewidth,trim={0cm 0cm 0cm 0cm}]{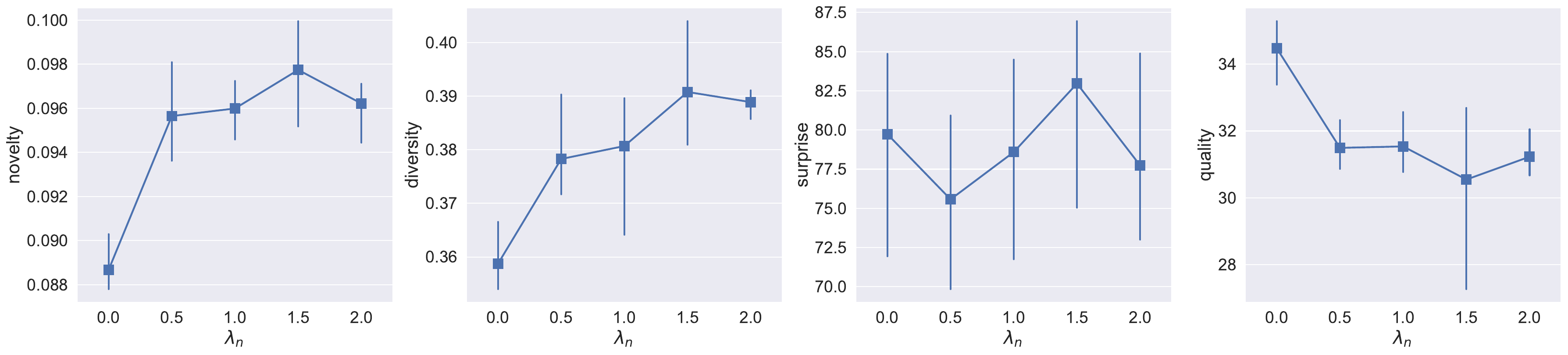}
    \caption{Effect of novelty injection weight on all creativity dimensions. Results are averaged across three seed runs.}
    \label{fig:lambda-results-novelty-int}
\end{figure*}

\begin{figure*}[h]
    \centering   
    \includegraphics[width=\linewidth,trim={0cm 0cm 0cm 0cm}]{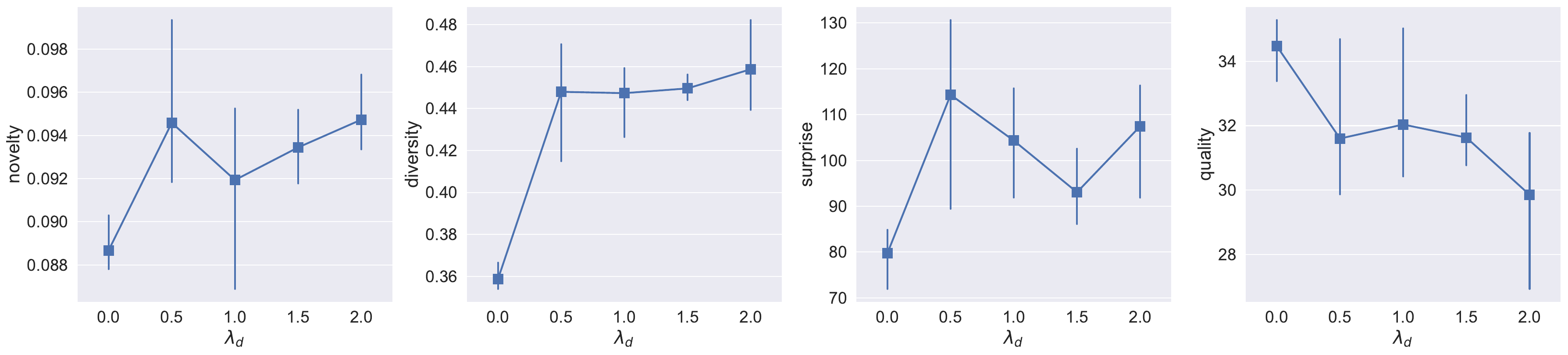}
    \caption{Effect of diversity injection weight on all creativity dimensions. Results are averaged across three seed runs.}
    \label{fig:lambda-results-diversity-int}
\end{figure*}

\begin{figure*}[h]
    \centering   
    \includegraphics[width=\linewidth,trim={0cm 0cm 0cm 0cm}]{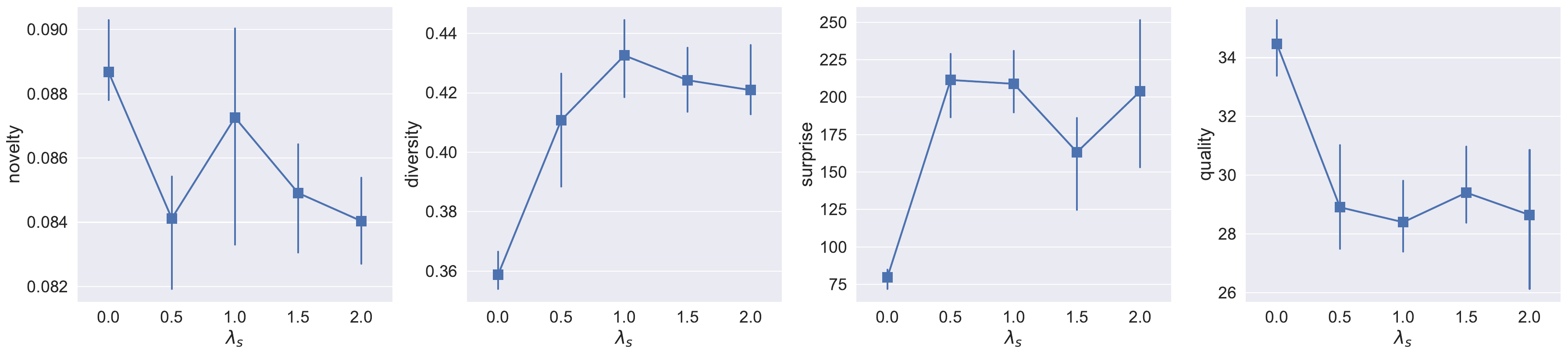}
    \caption{Effect of surprise injection weight on all creativity dimensions. Results are averaged across three seed runs.}
    \label{fig:lambda-results-surprise-int}
\end{figure*}

\section{Evaluation}
\label{sec:app-eval-details}
For each prompt in our held-out evaluation suite, we generate a total of $16$ responses for every model by sampling $4$ responses for each of the following four decoding setups that induce high randomness using various sampling techniques \cite{fan-etal-2018-hierarchical,holtzman2019curious}:
\begin{enumerate}
    \item  $temperature=0.7$, $top_p=0.95$
    \item  $temperature=0.9$, $top_p=0.99$
    \item  $temperature=0.7$, $top_k=50$
    \item  $temperature=0.8$, $top_p=0.97$
\end{enumerate}
Moreover, as the existing instruction-tuned LLMs tend to produce verbose outputs \cite{saito2023verbositybiaspreferencelabeling}, in order to minimize the length bias, we add further instructions in the prompt, constraining the output length in terms of the number of sentences and words. We compute the constraint values based on the median number of words and sentences of responses per task from our training dataset. Table \ref{tab:eval-prompts} lists an example evaluation prompt for each task. Table \ref{tab:model-gens} lists an example response from all models to a single prompt.

\begin{table*}[h]
    \centering
    \begin{tabular}{p{6cm} p{8cm}}
    \hline
        \textbf{Task} & \textbf{Prompt} \\
        \hline
        \textit{Real-Life Creative Problem Solving} & ``Come up with an original and creative solution for the following real-world problem: Clara, a junior pre-med student, is working part-time and taking a 15 hour credit load at school. ...<skipped>... Please limit your response to 4 sentences and at most 75 words.'' \\
        \hline
        \textit{Alternate Uses of Objects} & ``Come up with an original and creative use for the following object: rope. Please limit your response to 1 sentence and at most 17 words.'' \\
        \hline
        \textit{Design Solutions} & ``Come up with an original and creative solution to reduce the amount of litter in public spaces and promote waste reduction and recycling. Please limit your response to 2 sentences and at most 36 words.'' \\
        \hline
        \textit{Hypothesis Generation} & ``Come up with an original and creative scientific hypothesis for the following scenario: You notice that dogs seem to like one of your friends, but cats seem to like another friend. What hypotheses do you have about why that is? Please limit your response to 1 sentence and at most 22 words.'' \\
        \hline
        \textit{Metaphors} & ``Come up with an original and creative metaphoric equivalent for the concept described below: Stomata are tiny openings or pores found on the underside of a plant leaf. They are used for gas exchange, enabling the intake of carbon dioxide and release of oxygen.. Please limit your response to 1 sentence and at most 10 words.'' \\
        \hline
        \textit{Poems} & ``Come up with an original and creative poem about the following concept: choice. Please limit your response to 5 sentences and at most 150 words.'' \\
        \hline
        \textit{Sentence Completion} & ``Finish the sentence with an original and creative ending: When I got on the school bus....Please respond in one sentence.'' \\
      \hline
    \end{tabular}
    \caption{\MUSE{} dataset held-out item and task evaluation prompts.}
    \label{tab:eval-prompts}
\end{table*}

\subsection{Human Evaluation}
\label{sec:app-human-studies}
Since we have multiple model responses per prompt, instead of randomly choosing a response, for each prompt, we choose top 4 model responses measured by the overall automated creativity score which we define as the sum of normalized novelty, diversity, surprise and quality scores. This setup ensures that models are compared to each other with their best outputs. We recruited 15 participants on Prolific\footnote{https://www.prolific.com/} to complete the study, requiring that they reside in the U.S. and have an approval rating of at least 90\%. Ethics board approval was received from the Pennsylvania State University IRB for this study. We provided participants with a definition of creativity, and instructed them not to focus on the length or detail of the response when rating. 
Figure \ref{fig:creativity-human-instructions} lists the instructions given to raters for evaluating creativity. We additionally included a comprehension check where participants were quizzed about the task instructions, to help catch careless participants. Raters who failed this check were excluded from further analysis. All raters were compensated adequately with at least a minimum payment of 9\$ per hour. Final win rates are calculated for each response pair based on the majority vote across participants. The inter-rater agreement computed using Krippendorff's alpha was $0.463$, indicating a moderate agreement. 

\begin{table*}[h]
    \centering
    \begin{tabular}{c|c|c|c|c}
        \textbf{Model} & \textbf{novelty} & \textbf{diversity} & \textbf{surprise} & \textbf{quality} \\
         \hline
         \texttt{Llama-3.1-8B} & $0.05$ & $0.21$ & $46.28$ & $31.59$ \\
         \texttt{Llama-3.1-8B} (min-p) & $0.06$ & $0.22$ & $51.42$ & $31.31$ \\
         \CRPO{}\texttt{-nov} & $\mathbf{0.10}$ & $0.36$ & $71.75$ & $32.56$ \\
         \CRPO{}\texttt{-div} & $0.09$ & $\mathbf{0.46}$ & $115.73$ & $30.42$ \\
         \CRPO{}\texttt{-sur} & $0.08$ & $0.44$ & $\mathbf{230.91}$ & $27.39$ \\
         \CRPO{}\texttt{-nov-div-sur} & $0.09$ & $0.43$ & $120.07$ & $29.24$ \\
         \CRPO{}\texttt{-cre} & $0.09$ & $0.39$ & $96.92$ & $\mathbf{33.51}$ \\
    \end{tabular}
    \caption{Evaluation results comparing a creative decoding strategy min-p sampling, to our approach using \texttt{Llama-3.1-8B} as a base model. Best performance is \textbf{bolded} for each metric.}
    \label{tab:res-creative-decoding}
\end{table*}

\subsection{\textsc{NoveltyBench} Evaluation}
\label{sec:app-nbench-details}
\textsc{NoveltyBench} is a recently introduced benchmark to measure how well language models can generate novel and high-quality answers to user requests involving subjectivity, randomness, and creativity \cite{zhang2025noveltybench}. We use a 100-sample subset of their benchmark that is manually curated by the authors and contains four distinct categories where diversity and novelty are expected:
\begin{itemize}
    \item \textbf{Randomness:} prompts that involve randomizing over a set of options. Example: \textit{Roll a make-believe 20-sided die.}
    \item \textbf{Factual Knowledge:} prompts that request underspecified factual information, which allow many valid answers. Example: \textit{List a capital city in Africa.}
    \item \textbf{Creative Writing:} prompts that involve generating a creative form of text, including poetry, and story-writing. Example: \textit{Tell me a riddle.}
    \item \textbf{Subjectivity:} prompts that request subjective answers or opinions. Example: \textit{What’s the best car to get in 2023?}
\end{itemize} 

Additionally, the paper proposes new metrics to measure novelty and quality (i.e. utility) that are different than ours. To compute novelty, they propose a method that learns to partition the output space into equivalence classes from human annotations. Each class represents one unique generation that is roughly equivalent to the others in the same class and different from the generations in other classes. They consider a functional equivalence that defines two generations to be different if and only if a user who has seen one generation would likely benefit from seeing the other. To this end, the authors annotated 1,100 pairs of generations conditioned on prompts from \textsc{NoveltyBench} sampled from a diverse set of models. From these annotated pairs, they used 1,000 for training and fine-tuned a deberta-v3-large model \cite{he2023debertav3improvingdebertausing} to predict binary functional equivalence between two generations. With the equivalence classifier, they partition the output space into equivalence classes. Then they define the novelty as the $distinct_k$ metric that is the number of equivalence classes in a partition of k sample generations from a language model:
\begin{equation}
    distinct_k := |\{c_i | i \in [k]\}|
\end{equation}

To compute quality, they consider a model of user behavior that describes how users interact with and consume language model generations. They assume that the user has a patience level $p \in [0, 1]$: after
observing each additional generation, they have a probability $p$ of requesting an additional generation from the language model and observing the next generation, and a probability $1-p$ of stopping interacting with the model. Then they compute the quality of a sequence of generations as the cumulative utility:
\begin{equation}
    utility_k := \frac{1-p}{1-p^k} \sum_{i=1}^k {p^{i-1}} \cdot \mathbbm{1} [c_i \neq c_j, \forall j < i] \cdot u_i
\end{equation}

To compute the utility of individual generations, they also use the \texttt{Skywork-Reward-Gemma-2-27B-v0.2} \cite{liu2024skywork} model.

To benchmark our models, we follow their recommended setup for evaluation. In particular, we set the number of generations to 10 per model and the patience level to $0.8$ and use their trained classifier for output space partition.

\subsection{Additional Baselines}
\label{sec:app-additional-baselines}
\ME{We consider additional baseline comparisons using creative prompting and decoding strategies to further show the effectiveness of our approach.}

\subsubsection{Creative Prompting}
\ME{We choose an approach based on brainstorming and selecting \cite{SummersStay2023BrainstormTS} that has shown improvement on creativity scores in the Alternative Uses of Objects Task (AUT). In this method, an iterative prompting strategy is employed by asking the model first to generate a response, then brainstorm about the advantages and drawbacks of its response, and then evaluate the novelty and utility of its response, and propose different uses if needed. We follow the prompts suggested in \citet{SummersStay2023BrainstormTS} and evaluate one of our best-performing baselines, namely, GPT-4o model on the AUT task with a held-out set of 25 objects where for each object we generate 16 different responses using the same evaluation setup mentioned above. We also evaluate our \CRPO{} models on the same held-out test suite, but with standard single-turn prompting.}

\subsubsection{Creative Decoding}
\ME{Our main evaluation setup provides comparisons to popular decoding strategies such as top-p, top-k and temperature sampling that are often used to increase the creativity of language model outputs. However, these approaches are not designed to directly improve output creativity. Hence, to further show the effectiveness of our approach, we consider a recent decoding strategy called \textit{min-p sampling} that promotes output creativity directly \cite{nguyen2024turning}. We follow the paper’s recommended setup for best performance (min-p values 0.05, 0.1 with high-temperature values 1.0, 1.5) on creative tasks and evaluate the base model \texttt{Llama-3.1-8B} with this decoding setup on our multi-task evaluation suite.}

\begin{table*}[h]
    \centering
    \begin{tabular}{p{4cm} p{8cm}}
    \hline
        \textbf{Model} & \textbf{Response} \\
        \hline
        Human & ``played hide-and-seek by forming into different objects and trying not to be found.''\\
        \hline
        \texttt{Llama-3.1-8B-Instruct} & ``At a sleepover, we transformed our host's basement into a mystical 'Dreamscape' where we wove a tapestry of secrets, stories, and whispered promises that only the darkness could keep.'' \\
        \hline
        \texttt{Gemini-2.0-Flash} & ``At a sleepover we...built a pillow fort so magnificent, it accidentally opened a portal to a realm where dreams manifested as sentient, fluffy creatures demanding to be tickled.'' \\
        \hline
        \texttt{Claude-3.7-Sonnet} & ``At a sleepover we constructed an elaborate blanket fortress complete with secret passages, flashlight constellations on the ceiling, and an unspoken pact to guard each other's midnight confessions until the end of time.'' \\
        \hline
        \texttt{GPT-4o} & ``At a sleepover we transformed the living room into a magical fort of pillows and fairy lights, where we whispered secrets and told stories until the first light of dawn.'' \\
        \hline
        \texttt{SFT} & ``We played a game of Twister where we put out our phones to see who was the best twister.'' \\
        \hline
        \texttt{DPO} & ``conducted an experiment to see if a person's personality is changed with an altered state of consciousness.'' \\
        \hline
        \CRPO{}\texttt{-nov} & ``tried to break the record for most consecutive hours without speaking and we discovered we could all hear each other's thoughts.'' \\
        \hline
        \CRPO{}\texttt{-nov-qua} & ``Participated in an experiment where we tested the effects of sleep deprivation on the human mind.'' \\
        \hline
        \CRPO{}\texttt{-div} & ``construct a space shuttle that takes us to the moon and from there we can launch our dream rockets.'' \\
        \hline
        \CRPO{}\texttt{-div-qua} & ``recreated the conditions of a 19th century underground railroad and had to map out our escape to Canada.'' \\
        \hline
        \CRPO{}\texttt{-sur} & ``Operate on each other to implant a permanent adrenaline gland.'' \\
        \hline
        \CRPO{}\texttt{-sur-qua} & ``created an underwater laboratory within our inflatable pool to collect the evidence we found of alien life.'' \\
        \hline
        \CRPO{}\texttt{-qua} & ``began to master the art of telekinesis by competitively tossing each other's pillows across the room.'' \\
        \hline
        \CRPO{}\texttt{-nov-div-sur} & ``Built a rollercoaster out of air mattresses and then did a hot-wheel car-launch into the trenches and caught the crash on GoPro cameras.'' \\
        \hline
        \CRPO{}\texttt{-cre} & ``Created an experiment to see if our dreams could be manipulated and transfer to one another.'' \\
      \hline
    \end{tabular}
    \caption{Example model (and gold human) responses to the prompt ``At a sleepover we ...'' from the \textit{Sentence Completion} task.}
    \label{tab:model-gens}
\end{table*}

\begin{table*}[h]
    \centering
    \begin{tabular}{p{2cm} p{4cm} p{4cm} p{2cm} p{2cm}}
    \hline
        \textbf{Task} & \textbf{Description} & \textbf{Dataset Sources} & \textbf{\# prompts} & \textbf{\# samples} \\
        \hline
        \textit{Real-Life Creative Problem Solving} & Produce solutions for everyday problems & \cite{luchini2025automated, kapoor2024ties, saretzki2024investigation, perchtold2024development} & 28 & 33,340 \\
        \hline
        \textit{Alternate Titles Generation} & Produce alternative titles for widely known books or movies & \cite{agnoli2016estimating} & 6 & 2,986 \\
        \hline
        \textit{Question Asking} & Produce questions about everyday objects & \cite{raz2024bridging} & 6 & 3,566 \\
        \hline
        \textit{Poems} & Produce poems about everyday concepts & \cite{fan2023exploring, chaudhuri2025evaluating, he2022impact} & 15 & 2,580 \\
        \hline
        \textit{Design Solutions} & Produce solutions to real-world design problems & \cite{luchini2025automated, distefano2025evaluating} & 20 & 10,818 \\
        \hline
        \textit{Combining Objects} & Produce combinations of everyday objects to achieve a goal & \cite{weiss2023conceptualizing} & 13 & 4,494 \\
        \hline
        \textit{Plot Titles Generation} & Produce titles for story plots & \cite{weiss2023conceptualizing, goecke2024driving, weiss2024much} & 6 & 1,832 \\
        \hline
        \textit{Instances of Common Concepts} & Produce instances related to everyday adjectives & \cite{organisciak2023beyond} & 8 & 2,474 \\
        \hline
        \textit{Experiment Design} & Produce experiment designs to test scientific hypotheses & \cite{beaty2024scientific, goecke2024automated} & 7 & 4,893 \\
        \hline
        \textit{Associations} & Produce word associations & \cite{beaty2021automating} & 8 & 1,004 \\
        \hline
        \textit{Emotional Trials} & Produce feelings one might have in a given situation & \cite{weiss2023conceptualizing} & 6 & 732 \\
        \hline
        \textit{Invent Nicknames} & Produce nicknames for everyday concepts and objects & \cite{weiss2023conceptualizing} & 6 & 613 \\
        \hline
        \textit{Situation Redescription} & Produce redescriptions of negative situations into positive situations & \cite{weiss2023conceptualizing} & 6 & 826 \\
        \hline
        \textit{Alternate Uses of Objects Task} & Produce alternate uses for everyday objects & \cite{patterson2023multilingual, zielinska2023lost, organisciak2023beyond} & 211 & 88,155 \\
        \hline
        \textit{Stories} & Produce short stories from three word prompts & \cite{luchini2025automated, agnoli2016estimating, fan2023exploring, he2022impact} & 10 & 2,757 \\
      \hline
    \end{tabular}
    \caption{\MUSE{} dataset details broken down by task (Part 1).}
    \label{tab:muse-task-desc}
\end{table*}

\begin{table*}[h]
    \centering
    \begin{tabular}{p{2cm} p{4cm} p{4cm} p{2cm} p{2cm}}
    \hline
        \textbf{Task} & \textbf{Description} & \textbf{Dataset Sources} & \textbf{\# prompts} & \textbf{\# samples} \\
        \hline
        \textit{Malevolent Problems} & Produce ideas on how to take revenge on or sabotage a wrongdoer & \cite{perchtold2023functional, kapoor2024ties, perchtold2024development} & 36 & 16,536 \\
        \hline
        \textit{Metaphors} & Produce metaphors to describe scenarios & \cite{distefano2024automatic, yu2024exploring} & 136 & 13,210 \\
        \hline
        \textit{Essays} & Produce essays on a topic & \cite{cotter2016applicant} & 1 & 821 \\
        \hline
        \textit{Consequences} & Produce possible consequences to scenarios & \cite{weiss2024much, weiss2023conceptualizing, goecke2024driving} & 20 & 24,874 \\
        \hline
        \textit{Sentence Completion} & Produce endings to incomplete sentences & \cite{organisciak2023beyond} & 8 & 2,629 \\
        \hline
        \textit{Hypothesis Generation} & Produce scientific hypotheses for specific observations & \cite{beaty2024scientific, goecke2024automated} & 9 & 18,455 \\
        \hline
        \textit{Research Questions} & Produce research questions relating to scenarios & \cite{beaty2024scientific, goecke2024automated} & 5 & 5,161 \\
        \hline
        \textit{Composites} & Produce composite words from a prompt word & \cite{weiss2023conceptualizing} & 7 & 667 \\
        \hline
        \textit{Evoking Emotional Responses from People} & Produce ways to evoke emotional responses in people as a TV producer & \cite{weiss2023conceptualizing} & 6 & 789 \\
        \hline
        \textit{Emotions in Everyday Situations} & Produce emotions you might feel in response to everyday situations & \cite{weiss2023conceptualizing} & 6 & 818 \\
        \hline
    \end{tabular}
    \caption{\MUSE{} dataset details broken down by task (Part 2).}
    \label{tab:muse-task-desc2}
\end{table*}

\begin{table*}[h]
    \centering
    \begin{tabular}{p{2cm} p{4cm} p{4cm} p{4cm}}
    \hline
        \textbf{Task} & \textbf{Example prompt} & \textbf{Example low rating response} & \textbf{Example high rating response} \\
        \hline
        \textit{Real-Life Creative Problem Solving} & ``Becky is a college student who works part-time at Mark's Pizzeria. Mark, the owner of the restaurant, has treated Becky very well. He gave her a job that she needs to help pay her rent when no other business would employ her because she was arrested for shoplifting three years ago. Mark also lets Becky work around her school schedule, and has asked if she wants to be a shift manager in the summers. Becky's roommate Jim also works at the pizzeria, but Jim has been causing a lot of problems at work. He always avoids doing his job, treats customers rudely, and makes a lot of mistakes with orders. Jim recently began stealing food from the pizzeria. Two days ago the pizzeria was short- staffed, so Jim and Becky were the only employees left at closing time. Jim made 10 extra pizzas and took them home to a party he was hosting without paying for them. Becky feels like she needs to do something about Jim's behavior. However, Becky is hesitant to tell Mark about Jim because Jim is a good friend to Becky. Becky also needs Jim to have a job so he can pay his portion of their rent. Becky does not know what to do..'' & ``Morally the right thing for Becky to do would be to tell her boss. However, to be a good friend would to be not to tell on Jim. The only creative solution to this problem would to be to try and talk to Jim one on one. Give Jim the decision of whether or nt he wants Becky to inform their boss of what he has been doing. As a friend he should understand where Becky is coming from and want to take the strain off her.'' & ``Becky should first discuss this with Jim, and tell him that he needs to either pay for the pizzas or he needs to go to the boss, and admit what he has done.  He will get caught in the end because eventually the ingredients will be missed.  The boss may unerstand, and perhaps will allow him to work off the pizzas somehow.  Maybe he could help out cleaning up around the restaurant.  If Jim will not tell his boss Becky should tell him.  She wouldn't necessarily have to come right out and tell on her coworker she could come up with a way for the boss to catch him at it.  If he does it once Jim will more than likely do it again.  She could tell the boss to check on the inventory.  She could have other people who might have been at the party come tell her boss bout it.  If all of that fails, she should just tell Mark about Jim stealing the pizzas.'' \\
      \hline
    \end{tabular}
    \caption{\MUSE{} dataset examples (Part 1).}
    \label{tab:muse-task-ex}
\end{table*}

\begin{table*}[h]
    \centering
    \begin{tabular}{p{2cm} p{4cm} p{4cm} p{4cm}}
    \hline
        \textbf{Task} & \textbf{Example prompt} & \textbf{Example low rating response} & \textbf{Example high rating response} \\
    \hline
        \textit{Question Asking} & ``pencil'' & ``How big is it?'' & ``How many great ideas have started with a pencil?'' \\
        \hline
        \textit{Poems} & ``childhood'' & ``Twinkle, Twinkle little star....ect'' & ``Red Rover, Red Rover  Is my childhood over?  I don't feel quite grown up  I still laugh at "I CUP"  I play slide with my sister  and still call my fourth grade teacher "mister"  I suppose, even still, my childhood is over  even if I can still play red rover red rover'' \\
        \hline
        \textit{Design Solutions} & ``Develop as many design ideas as you can to reduce air pollution in cities.'' & ``Walk'' & ``use 3d printing as an innivating way of building houses as it reduces labour and'' \\
        \hline
        \textit{Combining Objects} & ``Paint sign'' & ``paper, ballpoint pen'' & ``beetroot juice, quark cheese'' \\
        \hline
        \textit{Plot Titles Generation} & ``Now spoke'' & ``A completely normal everyday life'' & ``VR glasses charger defective'' \\
        \hline
        \textit{Instances of Common Concepts} & ``soft'' & ``something that is not hard'' & ``a futuristic ball that turns really fuzzy and comfy at places it gets contact to'' \\
        \hline
        \textit{Experiment Design} & ``You think some animals have a sense of humor that humans don't usually understand. How could you test that hypothesis?'' & ``observe'' & ``tickle your dog to see how he acts when he's 'laughing.' then, observe your dog throughout the day and note when he is 'laughing.' you may begin to pick up on moments where he does things that are funny to him.'' \\
        \hline
        \textit{Associations} & ``expert'' & ``winner'' & ``ace'' \\
        \hline
        \textit{Emotional Trials} & ``You have a date tonight, and once again your dress didn't get ready in time at the laundry.'' & ``worried, afraid, sad'' & ``Anger, panic, anticipation'' \\
        \hline
        \textit{Invent Nicknames} & ``plate'' & ``porcelain'' & ``Shrunken UFO'' \\
      \hline
    \end{tabular}
    \caption{\MUSE{} dataset examples (Part 2).}
    \label{tab:muse-task-ex2}
\end{table*}

\begin{table*}[h]
    \centering
    \begin{tabular}{p{2cm} p{4cm} p{4cm} p{4cm}}
    \hline
        \textbf{Task} & \textbf{Example prompt} & \textbf{Example low rating response} & \textbf{Example high rating response} \\
        \hline
        \textit{Alternate Uses of Objects Task} & ``knife'' & ``weapon'' & ``make up "knife characters" and create a movie'' \\
        \hline
        \textit{Stories} & ``petrol-diesel-pump'' & ``I needed to fuel my car before we could start the long drive. I drove to the petrol station. i went to the pump  and fuel my car with diesel. new i was ready for the task ahead'' & ``Manly Merde was a truck driver looking for trouble.  He pulled into the Casino in the back where the drivers go.  He took a swig of whisky and walked to the petrol station, grabbed the pump and spurt diesel into the air like hydrocarbon fountain.  He let out a big belly laugh and screamed, "Let the revolution begin!"  And that is how the trucker wars started.'' \\
        \hline
        \textit{Malevolent Problems} & ``Your professor in class announces an award for the person who comes up with the best solution for a project. By chance, another student leaves their notebook behind in class. You read their ideas and believe that they are the best. You decide to turn them in as your own; however you know that if the other student submits the same solution, there will be a problem.'' & ``I will not do the above'' & ``render their notebook unreadable by dropping water at the last moment'' \\
        \hline
        \textit{Metaphors} & ``The hot tea is...'' & ``boiling'' & ``liquid fire'' \\
        \hline
        \textit{Consequences} & ``What would be the result if society no longer used money, and instead traded goods and services?'' & ``Banks would be unnecessary.'' & ``People (especially couples) would stop fighting so much about financial issues'' \\
        \hline
        \textit{Sentence Completion} & ``It started raining and...'' & ``I got wet'' & ``because I was covered in oil, I began to levitate, and all the witnesses called me the next coming of some sort of goddess.'' \\
        \hline
    \end{tabular}
    \caption{\MUSE{} dataset examples (Part 3).}
    \label{tab:muse-task-ex3}
\end{table*}

\begin{table*}[h]
    \centering
    \begin{tabular}{p{2cm} p{4cm} p{4cm} p{4cm}}
    \hline
        \textbf{Task} & \textbf{Example prompt} & \textbf{Example low rating response} & \textbf{Example high rating response} \\
    \hline
        \textit{Hypothesis Generation} & ``On a field trip, you drive past a massive field with hundreds of large holes visible as far as the eye can see. What hypotheses do you have about what purpose the holes may serve?'' & ``the holes resulted over time and nature'' & ``the holes are for animals giving birth.'' \\
        \hline
        \textit{Essays} & ``dream project'' & ``I don't really know what carreer path I want to follow.  I just want a job where I can help people and get a good pay check so I can support myt future endevors.  I want to do something that no one has ever done before in a way no one has ever seen.  I want to inspire a generation to work on a better future for everybody.  I guess what I really want is to be remembered as an icon.  i want to be someone that people look up to.'' & ``I want to go into forensic science when I gradutate. Therefore, my dream project is to discover the perfect device that can help solve every crime scene. This device would be able to analyize the crime scene and tell us exactly how many people died and how they died. It would then collect evidence samples such as blood. Next, it would use what the information it found at the crime scene to help make up questions the dectectives would ask the suspects. It would use it's technology to come up with questions that only the murderer could answer. Later on, back at the lab, it would help discover whose blood the samples belonged to. In the end, the only human power that would be needed was someone to arrest the convicted person and the people to help clean up the crime scene.'' \\
        \hline
    \end{tabular}
    \caption{\MUSE{} dataset examples (Part 4).}
    \label{tab:muse-task-ex4}
\end{table*}

\begin{table*}[h]
    \centering
    \begin{tabular}{p{2cm} p{4cm} p{4cm} p{4cm}}
    \hline
        \textbf{Task} & \textbf{Example prompt} & \textbf{Example low rating response} & \textbf{Example high rating response} \\
        \hline
        \textit{Situation Redescription} & ``You notice how your colleague first treats another employee very kindly and then shortly afterwards starts talking negatively behind his back'' & ``It would be nice if you were older'' & ``I'll talk to them. Then I'll have to work less'' \\
        \hline
        \textit{Alternate Titles Generation} & ``The Betrothed'' & ``renzo and lucia'' & ``Plague, Honor and Love in Baroque Brianza'' \\
        \hline
        \textit{Research Questions} & ``You travel to a jungle that contains no human life and is completely unknown to the scientific community. What scientific questions could you ask about this jungle?'' & ``How many people will come with me?'' & ``Do these species share a common characterisitic that humans don't have?'' \\
        \hline
        \textit{Composites} & ``jitters'' & ``Exam jitters'' & ``Easter bunny missing jitters'' \\
        \hline
        \textit{Evoking Emotional Responses from People} & ``Describe how you would make people look down on others'' & ``I will always scream loudly'' & ``I would divide the audience into two groups and give one group a rubber glove as headgear and the other group a tiara or crown made of real gold.' \\
        \hline
        \textit{Emotions in Everyday Situations} & ``You're at work. A glance at the clock tells you that you're about to finish work and start your long-awaited weekend.'' & ``I feel happy'' & ``I feel sorry for my desk chair, which is unused over the weekend and stands alone in the office.'' \\
        \hline
    \end{tabular}
    \caption{\MUSE{} dataset examples (Part 5).}
    \label{tab:muse-task-ex5}
\end{table*}

%% file: main.bbl
\begin{thebibliography}{120}
\providecommand{\natexlab}[1]{#1}

\bibitem[{Agnoli et~al.(2016)Agnoli, Corazza, and Runco}]{agnoli2016estimating}
Sergio Agnoli, Giovanni~E Corazza, and Mark~A Runco. 2016.
\newblock Estimating creativity with a multiple-measurement approach within scientific and artistic domains.
\newblock \emph{Creativity Research Journal}, 28(2):171--176.

\bibitem[{AI@Meta(2024)}]{llama3modelcard}
AI@Meta. 2024.
\newblock \href {https://github.com/meta-llama/llama3/blob/main/MODEL_CARD.md} {Llama 3 model card}.

\bibitem[{Anderson et~al.(2024)Anderson, Shah, and Kreminski}]{anderson2024homogenization}
Barrett~R Anderson, Jash~Hemant Shah, and Max Kreminski. 2024.
\newblock Homogenization effects of large language models on human creative ideation.
\newblock In \emph{Proceedings of the 16th conference on creativity \& cognition}, pages 413--425.

\bibitem[{Anthropic(2025)}]{claude37}
Anthropic. 2025.
\newblock \href {https://www.anthropic.com/news/claude-3-7-sonnet} {Claude 3.7 sonnet and claude code}.

\bibitem[{Barron(1955)}]{barron1955disposition}
Frank Barron. 1955.
\newblock The disposition toward originality.
\newblock \emph{The Journal of Abnormal and Social Psychology}, 51(3):478.

\bibitem[{Beaty et~al.(2024)Beaty, Cortes, Luchini, Patterson, Forthmann, Baker, Barbot, Hardiman, and Green}]{beaty2024scientific}
Roger Beaty, Robert~A Cortes, Simone Luchini, John~D Patterson, Boris Forthmann, Brendan~S Baker, Baptiste Barbot, Mariale Hardiman, and Adam Green. 2024.
\newblock The scientific creative thinking test (sctt): Reliability, validity, and automated scoring.
\newblock \emph{PsyArxiv Preprints}.

\bibitem[{Beaty and Johnson(2021)}]{beaty2021automating}
Roger~E Beaty and Dan~R Johnson. 2021.
\newblock Automating creativity assessment with semdis: An open platform for computing semantic distance.
\newblock \emph{Behavior research methods}, 53(2):757--780.

\bibitem[{Bellemare-Pepin et~al.(2024)Bellemare-Pepin, Lespinasse, Th{\"o}lke, Harel, Mathewson, Olson, Bengio, and Jerbi}]{bellemare2024divergent}
Antoine Bellemare-Pepin, Fran{\c{c}}ois Lespinasse, Philipp Th{\"o}lke, Yann Harel, Kory Mathewson, Jay~A Olson, Yoshua Bengio, and Karim Jerbi. 2024.
\newblock Divergent creativity in humans and large language models.
\newblock \emph{arXiv preprint arXiv:2405.13012}.

\bibitem[{Boden(2004)}]{boden2004creative}
Margaret~A Boden. 2004.
\newblock \emph{The creative mind: Myths and mechanisms}.
\newblock Routledge.

\bibitem[{Brown et~al.(2020)Brown, Mann, Ryder, Subbiah, Kaplan, Dhariwal, Neelakantan, Shyam, Sastry, Askell et~al.}]{brown2020language}
Tom Brown, Benjamin Mann, Nick Ryder, Melanie Subbiah, Jared~D Kaplan, Prafulla Dhariwal, Arvind Neelakantan, Pranav Shyam, Girish Sastry, Amanda Askell, and 1 others. 2020.
\newblock Language models are few-shot learners.
\newblock \emph{Advances in neural information processing systems}, 33:1877--1901.

\bibitem[{Bubeck et~al.(2023)Bubeck, Chandrasekaran, Eldan, Gehrke, Horvitz, Kamar, Lee, Lee, Li, Lundberg, Nori, Palangi, Ribeiro, and Zhang}]{bubeck2023sparks}
Sébastien Bubeck, Varun Chandrasekaran, Ronen Eldan, Johannes Gehrke, Eric Horvitz, Ece Kamar, Peter Lee, Yin~Tat Lee, Yuanzhi Li, Scott Lundberg, Harsha Nori, Hamid Palangi, Marco~Tulio Ribeiro, and Yi~Zhang. 2023.
\newblock \href {https://arxiv.org/abs/2303.12712} {Sparks of artificial general intelligence: Early experiments with gpt-4}.
\newblock \emph{Preprint}, arXiv:2303.12712.

\bibitem[{Bunescu and Uduehi(2022)}]{Bunescu2022DistributionBasedMO}
Razvan~C. Bunescu and Oseremen~O. Uduehi. 2022.
\newblock \href {https://api.semanticscholar.org/CorpusID:256461155} {Distribution-based measures of surprise for creative language: Experiments with humor and metaphor}.
\newblock \emph{Proceedings of the 3rd Workshop on Figurative Language Processing (FLP)}.

\bibitem[{Chakrabarty et~al.(2024)Chakrabarty, Laban, Agarwal, Muresan, and Wu}]{chakrabarty2024art}
Tuhin Chakrabarty, Philippe Laban, Divyansh Agarwal, Smaranda Muresan, and Chien-Sheng Wu. 2024.
\newblock Art or artifice? large language models and the false promise of creativity.
\newblock In \emph{Proceedings of the 2024 CHI Conference on Human Factors in Computing Systems}, pages 1--34.

\bibitem[{Chaudhuri et~al.(2025)Chaudhuri, Pickering, and Bhattacharya}]{chaudhuri2025evaluating}
Soma Chaudhuri, Alan Pickering, and Joydeep Bhattacharya. 2025.
\newblock Evaluating poetry: Navigating the divide between aesthetical and creativity judgments.
\newblock \emph{The Journal of Creative Behavior}, 59(1):e683.

\bibitem[{Chen et~al.(2024)Chen, Zhang, Wang, and Wu}]{chen2024weak}
Qi~Chen, Bowen Zhang, Gang Wang, and Qi~Wu. 2024.
\newblock Weak-eval-strong: Evaluating and eliciting lateral thinking of llms with situation puzzles.
\newblock \emph{arXiv preprint arXiv:2410.06733}.

\bibitem[{Chung et~al.(2023)Chung, Kamar, and Amershi}]{chung2023increasing}
John Joon~Young Chung, Ece Kamar, and Saleema Amershi. 2023.
\newblock Increasing diversity while maintaining accuracy: Text data generation with large language models and human interventions.
\newblock \emph{arXiv preprint arXiv:2306.04140}.

\bibitem[{Chung et~al.(2025)Chung, Padmakumar, Roemmele, Sun, and Kreminski}]{chung2025modifying}
John Joon~Young Chung, Vishakh Padmakumar, Melissa Roemmele, Yuqian Sun, and Max Kreminski. 2025.
\newblock Modifying large language model post-training for diverse creative writing.
\newblock \emph{arXiv preprint arXiv:2503.17126}.

\bibitem[{Cotter et~al.(2016)Cotter, Pretz, and Kaufman}]{cotter2016applicant}
Katherine~N Cotter, Jean~E Pretz, and James~C Kaufman. 2016.
\newblock Applicant extracurricular involvement predicts creativity better than traditional admissions factors.
\newblock \emph{Psychology of Aesthetics, Creativity, and the Arts}, 10(1):2.

\bibitem[{Cseh and Jeffries(2019)}]{cseh2019scattered}
Genevieve~M Cseh and Karl~K Jeffries. 2019.
\newblock A scattered cat: A critical evaluation of the consensual assessment technique for creativity research.
\newblock \emph{Psychology of Aesthetics, Creativity, and the Arts}, 13(2):159.

\bibitem[{de~Chantal et~al.(2025{\natexlab{a}})de~Chantal, Beaty, Laverghetta, Pronchick, Patterson, Organisciak, Potega~vel Zabik, Barbot, and Karwowski}]{deartificial}
Pier~Luc de~Chantal, Roger Beaty, Antonio Laverghetta, Jimmy Pronchick, John Patterson, Peter Organisciak, Katarzyna Potega~vel Zabik, Baptiste Barbot, and Maciej Karwowski. 2025{\natexlab{a}}.
\newblock Artificial intelligence enhances human creativity through real-time evaluative feedback.

\bibitem[{de~Chantal et~al.(2025{\natexlab{b}})de~Chantal, Beaty, Laverghetta, Pronchick, Patterson, Organisciak, vel Zabik, Barbot, and Karwowski}]{de2025artificial}
Pier~Luc de~Chantal, Roger Beaty, Antonio Laverghetta, Jimmy Pronchick, John Patterson, Peter Organisciak, Katarzyna~Potega vel Zabik, Baptiste Barbot, and Maciej Karwowski. 2025{\natexlab{b}}.
\newblock Artificial intelligence enhances human creativity through real-time evaluative feedback.

\bibitem[{De~Chantal and Organisciak(2023)}]{de2023automated}
Pier-Luc De~Chantal and Peter Organisciak. 2023.
\newblock Automated feedback and creativity: On the role of metacognitive monitoring in divergent thinking.
\newblock \emph{Psychology of Aesthetics, Creativity, and the Arts}.

\bibitem[{Deng et~al.(2025)Deng, Zhong, Ai, Feng, Wang, and He}]{deng2025less}
Xun Deng, Han Zhong, Rui Ai, Fuli Feng, Zheng Wang, and Xiangnan He. 2025.
\newblock Less is more: Improving llm alignment via preference data selection.
\newblock \emph{arXiv preprint arXiv:2502.14560}.

\bibitem[{DiStefano et~al.(2024)DiStefano, Patterson, and Beaty}]{distefano2024automatic}
Paul~V DiStefano, John~D Patterson, and Roger~E Beaty. 2024.
\newblock Automatic scoring of metaphor creativity with large language models.
\newblock \emph{Creativity Research Journal}, pages 1--15.

\bibitem[{DiStefano et~al.(2025)DiStefano, Zeitlen, Rafner, de~Chantal, Peng, Miller, and Beaty}]{distefano2025evaluating}
Paul~V DiStefano, Daniel Zeitlen, Janet Rafner, Pier-Luc de~Chantal, Aoran Peng, Scarlett Miller, and Roger Beaty. 2025.
\newblock Evaluating ai’s ideas: The role of individual creativity and expertise in human-ai co-creativity.

\bibitem[{Dunbar and Forster(2009)}]{dunbar2009creativity}
Kevin Dunbar and Eve Forster. 2009.
\newblock Creativity evaluation through latent semantic analysis.
\newblock In \emph{Proceedings of the Annual Meeting of the Cognitive Science Society}, volume~31.

\bibitem[{Fan et~al.(2018)Fan, Lewis, and Dauphin}]{fan-etal-2018-hierarchical}
Angela Fan, Mike Lewis, and Yann Dauphin. 2018.
\newblock \href {https://doi.org/10.18653/v1/P18-1082} {Hierarchical neural story generation}.
\newblock In \emph{Proceedings of the 56th Annual Meeting of the Association for Computational Linguistics (Volume 1: Long Papers)}, pages 889--898, Melbourne, Australia. Association for Computational Linguistics.

\bibitem[{Fan et~al.(2023)Fan, Zhuang, Wang, Zhang, Liu, Gu, and Qiu}]{fan2023exploring}
Li~Fan, Kaixiang Zhuang, Xueyang Wang, Jingyi Zhang, Cheng Liu, Jing Gu, and Jiang Qiu. 2023.
\newblock Exploring the behavioral and neural correlates of semantic distance in creative writing.
\newblock \emph{Psychophysiology}, 60(5):e14239.

\bibitem[{Forthmann et~al.(2025)Forthmann, Goecke, and Beaty}]{forthmann2025planning}
Boris Forthmann, Benjamin Goecke, and Roger~E Beaty. 2025.
\newblock Planning missing data designs for human ratings in creativity research: A practical guide.
\newblock \emph{Creativity Research Journal}, 37(1):167--178.

\bibitem[{Forthmann et~al.(2017)Forthmann, Holling, Zandi, Gerwig, {\c{C}}elik, Storme, and Lubart}]{forthmann2017missing}
Boris Forthmann, Heinz Holling, Nima Zandi, Anne Gerwig, P{\i}nar {\c{C}}elik, Martin Storme, and Todd Lubart. 2017.
\newblock Missing creativity: The effect of cognitive workload on rater (dis-) agreement in subjective divergent-thinking scores.
\newblock \emph{Thinking Skills and Creativity}, 23:129--139.

\bibitem[{Forum(2025)}]{World_Economic_Forum_2025a}
World~Economic Forum. 2025.
\newblock \href {https://reports.weforum.org/docs/WEF_Future_of_Jobs_Report_2025.pdf} {Future of jobs report}.

\bibitem[{Franceschelli and Musolesi(2024)}]{franceschelli2024creative}
Giorgio Franceschelli and Mirco Musolesi. 2024.
\newblock Creative beam search: Llm-as-a-judge for improving response generation.
\newblock \emph{arXiv preprint arXiv:2405.00099}.

\bibitem[{Gao et~al.(2024)Gao, Song, Miao, Cai, Yang, Chen, Hu, Xu, Dong, Zheng, Xiao, Zhang, Zan, Lu, Yu, Liu, Cui, Yang, Sha, Wang, Sui, Wang, Liu, and Chang}]{gao2024unifiedviewpreferencelearning}
Bofei Gao, Feifan Song, Yibo Miao, Zefan Cai, Zhe Yang, Liang Chen, Helan Hu, Runxin Xu, Qingxiu Dong, Ce~Zheng, Wen Xiao, Ge~Zhang, Daoguang Zan, Keming Lu, Bowen Yu, Dayiheng Liu, Zeyu Cui, Jian Yang, Lei Sha, and 5 others. 2024.
\newblock \href {https://arxiv.org/abs/2409.02795} {Towards a unified view of preference learning for large language models: A survey}.
\newblock \emph{Preprint}, arXiv:2409.02795.

\bibitem[{Gilhooly(2024)}]{gilhooly2024ai}
Ken Gilhooly. 2024.
\newblock Ai vs humans in the aut: Simulations to llms.
\newblock \emph{Journal of Creativity}, 34(1):100071.

\bibitem[{Goecke et~al.(2024{\natexlab{a}})Goecke, DiStefano, Aschauer, Haim, Beaty, and Forthmann}]{goecke2024automated}
Benjamin Goecke, Paul~V DiStefano, Wolfgang Aschauer, Kurt Haim, Roger Beaty, and Boris Forthmann. 2024{\natexlab{a}}.
\newblock Automated scoring of scientific creativity in german.
\newblock \emph{The Journal of Creative Behavior}, 58(3):321--327.

\bibitem[{Goecke et~al.(2024{\natexlab{b}})Goecke, Weiss, and Wilhelm}]{goecke2024driving}
Benjamin Goecke, Selina Weiss, and Oliver Wilhelm. 2024{\natexlab{b}}.
\newblock Driving factors of individual differences in broad retrieval ability: Gr is more than the sum of its parts.
\newblock \emph{Journal of Experimental Psychology: Learning, Memory, and Cognition}.

\bibitem[{G{\'o}es et~al.(2023)G{\'o}es, Sawicki, Grzes, Volpe, and Watson}]{GesPushingGC}
Fabr{\'i}cio G{\'o}es, Piotr Sawicki, Marek Grzes, Marco Volpe, and Jacob Watson. 2023.
\newblock \href {https://api.semanticscholar.org/CorpusID:260844252} {Pushing gpt's creativity to its limits: Alternative uses and torrance tests}.
\newblock In \emph{ICCC}.

\bibitem[{Google(2024{\natexlab{a}})}]{gemmateam2024gemma2improvingopen}
Google. 2024{\natexlab{a}}.
\newblock \href {https://arxiv.org/abs/2408.00118} {Gemma 2: Improving open language models at a practical size}.

\bibitem[{Google(2024{\natexlab{b}})}]{gemini20}
Google. 2024{\natexlab{b}}.
\newblock \href {https://blog.google/technology/google-deepmind/google-gemini-ai-update-december-2024/#ceo-message} {Introducing gemini 2.0: our new ai model for the agentic era}.

\bibitem[{Grace and Maher(2016)}]{grace2016surprise}
Kazjon Grace and Mary~Lou Maher. 2016.
\newblock Surprise-triggered reformulation of design goals.
\newblock In \emph{Proceedings of the AAAI Conference on Artificial Intelligence}, volume~30.

\bibitem[{Gugger et~al.(2022)Gugger, Debut, Wolf, Schmid, Mueller, Mangrulkar, Sun, and Bossan}]{accelerate}
Sylvain Gugger, Lysandre Debut, Thomas Wolf, Philipp Schmid, Zachary Mueller, Sourab Mangrulkar, Marc Sun, and Benjamin Bossan. 2022.
\newblock Accelerate: Training and inference at scale made simple, efficient and adaptable.
\newblock \url{https://github.com/huggingface/accelerate}.

\bibitem[{Guilford(1967)}]{guilford1967nature}
J.P. Guilford. 1967.
\newblock \href {https://books.google.ch/books?id=T-ZJAAAAMAAJ} {\emph{The Nature of Human Intelligence}}.
\newblock McGraw-Hill series in psychology. McGraw-Hill.

\bibitem[{Haase et~al.(2025)Haase, Hanel, and Pokutta}]{haase2025sdatmultilingualgenaidrivenframework}
Jennifer Haase, Paul H.~P. Hanel, and Sebastian Pokutta. 2025.
\newblock \href {https://arxiv.org/abs/2505.09068} {S-dat: A multilingual, genai-driven framework for automated divergent thinking assessment}.
\newblock \emph{Preprint}, arXiv:2505.09068.

\bibitem[{Harbinson and Haarman(2014)}]{harbinson2014automated}
J~Harbinson and Henk Haarman. 2014.
\newblock Automated scoring of originality using semantic representations.
\newblock In \emph{Proceedings of the Annual Meeting of the Cognitive Science Society}, volume~36.

\bibitem[{Hayati et~al.(2023)Hayati, Lee, Rajagopal, and Kang}]{hayati2023far}
Shirley~Anugrah Hayati, Minhwa Lee, Dheeraj Rajagopal, and Dongyeop Kang. 2023.
\newblock How far can we extract diverse perspectives from large language models?
\newblock \emph{arXiv preprint arXiv:2311.09799}.

\bibitem[{He et~al.(2023)He, Gao, and Chen}]{he2023debertav3improvingdebertausing}
Pengcheng He, Jianfeng Gao, and Weizhu Chen. 2023.
\newblock \href {https://arxiv.org/abs/2111.09543} {Debertav3: Improving deberta using electra-style pre-training with gradient-disentangled embedding sharing}.
\newblock \emph{Preprint}, arXiv:2111.09543.

\bibitem[{He et~al.(2022)He, Zhuang, Liu, Ding, Wang, Fu, Qiu, and Chen}]{he2022impact}
Ruizhi He, Kaixiang Zhuang, Lijun Liu, Ke~Ding, Xi~Wang, Lei Fu, Jiang Qiu, and Qunlin Chen. 2022.
\newblock The impact of knowledge on poetry composition: An fmri investigation.
\newblock \emph{Brain and language}, 235:105202.

\bibitem[{Holtzman et~al.(2019)Holtzman, Buys, Du, Forbes, and Choi}]{holtzman2019curious}
Ari Holtzman, Jan Buys, Li~Du, Maxwell Forbes, and Yejin Choi. 2019.
\newblock The curious case of neural text degeneration.
\newblock \emph{arXiv preprint arXiv:1904.09751}.

\bibitem[{Hu et~al.(2022)Hu, Shen, Wallis, Allen-Zhu, Li, Wang, Wang, Chen et~al.}]{hu2022lora}
Edward~J Hu, Yelong Shen, Phillip Wallis, Zeyuan Allen-Zhu, Yuanzhi Li, Shean Wang, Lu~Wang, Weizhu Chen, and 1 others. 2022.
\newblock Lora: Low-rank adaptation of large language models.
\newblock \emph{ICLR}, 1(2):3.

\bibitem[{Huang et~al.(2024)Huang, Ma, Li, Huang, Zou, Zhang, and Zheng}]{huang-etal-2024-lateval}
Shulin Huang, Shirong Ma, Yinghui Li, Mengzuo Huang, Wuhe Zou, Weidong Zhang, and Haitao Zheng. 2024.
\newblock \href {https://aclanthology.org/2024.lrec-main.889/} {{L}at{E}val: An interactive {LLM}s evaluation benchmark with incomplete information from lateral thinking puzzles}.
\newblock In \emph{Proceedings of the 2024 Joint International Conference on Computational Linguistics, Language Resources and Evaluation (LREC-COLING 2024)}, pages 10186--10197, Torino, Italia. ELRA and ICCL.

\bibitem[{Ismayilzada et~al.(2024{\natexlab{a}})Ismayilzada, Paul, Bosselut, and van~der Plas}]{ismayilzada2024creativity}
Mete Ismayilzada, Debjit Paul, Antoine Bosselut, and Lonneke van~der Plas. 2024{\natexlab{a}}.
\newblock Creativity in ai: Progresses and challenges.
\newblock \emph{arXiv preprint arXiv:2410.17218}.

\bibitem[{Ismayilzada et~al.(2024{\natexlab{b}})Ismayilzada, Stevenson, and van~der Plas}]{ismayilzada2024evaluating}
Mete Ismayilzada, Claire Stevenson, and Lonneke van~der Plas. 2024{\natexlab{b}}.
\newblock Evaluating creative short story generation in humans and large language models.
\newblock In \emph{Proceedings of ICCC 2025}.

\bibitem[{Jiang et~al.(2023)Jiang, Sablayrolles, Mensch, Bamford, Chaplot, de~las Casas, Bressand, Lengyel, Lample, Saulnier, Lavaud, Lachaux, Stock, Scao, Lavril, Wang, Lacroix, and Sayed}]{jiang2023mistral7b}
Albert~Q. Jiang, Alexandre Sablayrolles, Arthur Mensch, Chris Bamford, Devendra~Singh Chaplot, Diego de~las Casas, Florian Bressand, Gianna Lengyel, Guillaume Lample, Lucile Saulnier, Lélio~Renard Lavaud, Marie-Anne Lachaux, Pierre Stock, Teven~Le Scao, Thibaut Lavril, Thomas Wang, Timothée Lacroix, and William~El Sayed. 2023.
\newblock \href {https://arxiv.org/abs/2310.06825} {Mistral 7b}.
\newblock \emph{Preprint}, arXiv:2310.06825.

\bibitem[{Johnson et~al.(2021)Johnson, Cuthbert, and Tynan}]{johnson2021neglect}
Dan~R Johnson, Andrew~S Cuthbert, and Mara~E Tynan. 2021.
\newblock The neglect of idea diversity in creative idea generation and evaluation.
\newblock \emph{Psychology of Aesthetics, Creativity, and the Arts}, 15(1):125.

\bibitem[{Johnson et~al.(2022)Johnson, Kaufman, Baker, Patterson, Barbot, Green, van Hell, Kennedy, Sullivan, Taylor, Ward, and Beaty}]{Johnson2022DivergentSI}
Dan~Richard Johnson, J.~Kaufman, Brendan~S. Baker, John~D. Patterson, Baptiste Barbot, Adam~E. Green, Janet~G. van Hell, Evan~S. Kennedy, Grace~F Sullivan, Christa~L. Taylor, Thomas Ward, and Roger~E. Beaty. 2022.
\newblock \href {https://api.semanticscholar.org/CorpusID:252969336} {Divergent semantic integration (dsi): Extracting creativity from narratives with distributional semantic modeling}.
\newblock \emph{Behavior Research Methods}, 55:3726 -- 3759.

\bibitem[{Kapoor et~al.(2024)Kapoor, Mahadeshwar, Rezaei, Reiter-Palmon, and Kaufman}]{kapoor2024ties}
Hansika Kapoor, Hreem Mahadeshwar, Sarah Rezaei, Roni Reiter-Palmon, and James~C Kaufman. 2024.
\newblock The ties that bind: Low morals, high deception, and dark creativity.
\newblock \emph{Creativity Research Journal}, pages 1--20.

\bibitem[{Karampiperis et~al.(2014)Karampiperis, Koukourikos, and Koliopoulou}]{Karampiperis2014}
Pythagoras Karampiperis, Antonis Koukourikos, and Evangelia Koliopoulou. 2014.
\newblock \href {https://doi.org/10.1109/ICALT.2014.150} {Towards machines for measuring creativity: The use of computational tools in storytelling activities}.
\newblock In \emph{2014 IEEE 14th International Conference on Advanced Learning Technologies}, pages 508--512.

\bibitem[{Kirk et~al.(2023)Kirk, Mediratta, Nalmpantis, Luketina, Hambro, Grefenstette, and Raileanu}]{kirk2023understanding}
Robert Kirk, Ishita Mediratta, Christoforos Nalmpantis, Jelena Luketina, Eric Hambro, Edward Grefenstette, and Roberta Raileanu. 2023.
\newblock Understanding the effects of rlhf on llm generalisation and diversity.
\newblock \emph{arXiv preprint arXiv:2310.06452}.

\bibitem[{Koivisto and Grassini(2023)}]{koivisto2023best}
Mika Koivisto and Simone Grassini. 2023.
\newblock Best humans still outperform artificial intelligence in a creative divergent thinking task.
\newblock \emph{Scientific reports}, 13(1):13601.

\bibitem[{Kuznetsova et~al.(2013)Kuznetsova, Chen, and Choi}]{Kuznetsova2013UnderstandingAQ}
Polina Kuznetsova, Jianfu Chen, and Yejin Choi. 2013.
\newblock \href {https://api.semanticscholar.org/CorpusID:7782325} {Understanding and quantifying creativity in lexical composition}.
\newblock In \emph{Conference on Empirical Methods in Natural Language Processing}.

\bibitem[{Lambert et~al.(2024)Lambert, Pyatkin, Morrison, Miranda, Lin, Chandu, Dziri, Kumar, Zick, Choi et~al.}]{lambert2024rewardbench}
Nathan Lambert, Valentina Pyatkin, Jacob Morrison, LJ~Miranda, Bill~Yuchen Lin, Khyathi Chandu, Nouha Dziri, Sachin Kumar, Tom Zick, Yejin Choi, and 1 others. 2024.
\newblock Rewardbench: Evaluating reward models for language modeling.
\newblock \emph{arXiv preprint arXiv:2403.13787}.

\bibitem[{Lanchantin et~al.(2025)Lanchantin, Chen, Dhuliawala, Yu, Weston, Sukhbaatar, and Kulikov}]{lanchantin2025diverse}
Jack Lanchantin, Angelica Chen, Shehzaad Dhuliawala, Ping Yu, Jason Weston, Sainbayar Sukhbaatar, and Ilia Kulikov. 2025.
\newblock Diverse preference optimization.
\newblock \emph{arXiv preprint arXiv:2501.18101}.

\bibitem[{Liu et~al.(2024)Liu, Zeng, Liu, Yan, He, Wang, Yan, Liu, and Zhou}]{liu2024skywork}
Chris~Yuhao Liu, Liang Zeng, Jiacai Liu, Rui Yan, Jujie He, Chaojie Wang, Shuicheng Yan, Yang Liu, and Yahui Zhou. 2024.
\newblock Skywork-reward: Bag of tricks for reward modeling in llms.
\newblock \emph{arXiv preprint arXiv:2410.18451}.

\bibitem[{Loi and van~der Plas(2020)}]{loi2020blindspot}
Michele Loi and Lonneke van~der Plas. 2020.
\newblock A blindspot of ai ethics: anti-fragility in statistical prediction.
\newblock In \emph{Proceedings of the SwissText 2020}.

\bibitem[{Loi et~al.(2020)Loi, Vigan{\`o}, and van~der Plas}]{loi2020societal}
Michele Loi, Eleonora Vigan{\`o}, and Lonneke van~der Plas. 2020.
\newblock The societal and ethical relevance of computational creativity.
\newblock In \emph{Proceedings of the ICCC 2020}.

\bibitem[{Lu et~al.(2024)Lu, Sclar, Hallinan, Mireshghallah, Liu, Han, Ettinger, Jiang, Chandu, Dziri et~al.}]{lu2024ai}
Ximing Lu, Melanie Sclar, Skyler Hallinan, Niloofar Mireshghallah, Jiacheng Liu, Seungju Han, Allyson Ettinger, Liwei Jiang, Khyathi Chandu, Nouha Dziri, and 1 others. 2024.
\newblock Ai as humanity's salieri: Quantifying linguistic creativity of language models via systematic attribution of machine text against web text.
\newblock \emph{arXiv preprint arXiv:2410.04265}.

\bibitem[{Luchini et~al.(2025)Luchini, Maliakkal, DiStefano, Laverghetta~Jr, Patterson, Beaty, and Reiter-Palmon}]{luchini2025automated}
Simone~A Luchini, Nadine~T Maliakkal, Paul~V DiStefano, Antonio Laverghetta~Jr, John~D Patterson, Roger~E Beaty, and Roni Reiter-Palmon. 2025.
\newblock Automated scoring of creative problem solving with large language models: A comparison of originality and quality ratings.
\newblock \emph{Psychology of Aesthetics, Creativity, and the Arts}.

\bibitem[{Maher(2010)}]{maher2010evaluating}
Mary~Lou Maher. 2010.
\newblock Evaluating creativity in humans, computers, and collectively intelligent systems.
\newblock In \emph{Proceedings of the 1st DESIRE Network Conference on Creativity and Innovation in Design}, pages 22--28.

\bibitem[{Mehrotra et~al.(2024)Mehrotra, Parab, and Gulwani}]{mehrotra2024enhancing}
Pronita Mehrotra, Aishni Parab, and Sumit Gulwani. 2024.
\newblock Enhancing creativity in large language models through associative thinking strategies.
\newblock \emph{arXiv preprint arXiv:2405.06715}.

\bibitem[{Meister et~al.(2023)Meister, Pimentel, Wiher, and Cotterell}]{meister-etal-2023-locally}
Clara Meister, Tiago Pimentel, Gian Wiher, and Ryan Cotterell. 2023.
\newblock \href {https://doi.org/10.1162/tacl_a_00536} {Locally typical sampling}.
\newblock \emph{Transactions of the Association for Computational Linguistics}, 11:102--121.

\bibitem[{Modirshanechi et~al.(2022)Modirshanechi, Brea, and Gerstner}]{MODIRSHANECHI2022102712}
Alireza Modirshanechi, Johanni Brea, and Wulfram Gerstner. 2022.
\newblock \href {https://doi.org/10.1016/j.jmp.2022.102712} {A taxonomy of surprise definitions}.
\newblock \emph{Journal of Mathematical Psychology}, 110:102712.

\bibitem[{Myszkowski and Storme(2019)}]{myszkowski2019judge}
Nils Myszkowski and Martin Storme. 2019.
\newblock Judge response theory? a call to upgrade our psychometrical account of creativity judgments.
\newblock \emph{Psychology of Aesthetics, Creativity, and the Arts}, 13(2):167.

\bibitem[{Nair et~al.(2024)Nair, Gizzi, and Sinapov}]{nair-etal-2024-creative}
Lakshmi Nair, Evana Gizzi, and Jivko Sinapov. 2024.
\newblock \href {https://doi.org/10.18653/v1/2024.findings-emnlp.700} {Creative problem solving in large language and vision models - what would it take?}
\newblock In \emph{Findings of the Association for Computational Linguistics: EMNLP 2024}, pages 11978--11994, Miami, Florida, USA. Association for Computational Linguistics.

\bibitem[{Nguyen et~al.(2024)Nguyen, Baker, Neo, Roush, Kirsch, and Shwartz-Ziv}]{nguyen2024turning}
Minh Nguyen, Andrew Baker, Clement Neo, Allen Roush, Andreas Kirsch, and Ravid Shwartz-Ziv. 2024.
\newblock Turning up the heat: Min-p sampling for creative and coherent llm outputs.
\newblock \emph{arXiv preprint arXiv:2407.01082}.

\bibitem[{OpenAI(2024)}]{openai2024gpt4ocard}
OpenAI. 2024.
\newblock \href {https://arxiv.org/abs/2410.21276} {Gpt-4o system card}.
\newblock \emph{Preprint}, arXiv:2410.21276.

\bibitem[{Organisciak et~al.(2023)Organisciak, Acar, Dumas, and Berthiaume}]{organisciak2023beyond}
Peter Organisciak, Selcuk Acar, Denis Dumas, and Kelly Berthiaume. 2023.
\newblock Beyond semantic distance: Automated scoring of divergent thinking greatly improves with large language models.
\newblock \emph{Thinking Skills and Creativity}, 49:101356.

\bibitem[{Ouyang et~al.(2022)Ouyang, Wu, Jiang, Almeida, Wainwright, Mishkin, Zhang, Agarwal, Slama, Ray et~al.}]{ouyang2022training}
Long Ouyang, Jeffrey Wu, Xu~Jiang, Diogo Almeida, Carroll Wainwright, Pamela Mishkin, Chong Zhang, Sandhini Agarwal, Katarina Slama, Alex Ray, and 1 others. 2022.
\newblock Training language models to follow instructions with human feedback.
\newblock \emph{Advances in neural information processing systems}, 35:27730--27744.

\bibitem[{O’Mahony et~al.(2024)O’Mahony, Grinsztajn, Schoelkopf, and Biderman}]{o2024attributing}
Laura O’Mahony, Leo Grinsztajn, Hailey Schoelkopf, and Stella Biderman. 2024.
\newblock Attributing mode collapse in the fine-tuning of large language models.
\newblock In \emph{ICLR 2024 Workshop on Mathematical and Empirical Understanding of Foundation Models}.

\bibitem[{Padmakumar and He(2023)}]{padmakumar2023does}
Vishakh Padmakumar and He~He. 2023.
\newblock Does writing with language models reduce content diversity?
\newblock \emph{arXiv preprint arXiv:2309.05196}.

\bibitem[{Patterson et~al.(2023)Patterson, Merseal, Johnson, Agnoli, Baas, Baker, Barbot, Benedek, Borhani, Chen et~al.}]{patterson2023multilingual}
John~D Patterson, Hannah~M Merseal, Dan~R Johnson, Sergio Agnoli, Matthijs Baas, Brendan~S Baker, Baptiste Barbot, Mathias Benedek, Khatereh Borhani, Qunlin Chen, and 1 others. 2023.
\newblock Multilingual semantic distance: Automatic verbal creativity assessment in many languages.
\newblock \emph{Psychology of Aesthetics, Creativity, and the Arts}, 17(4):495.

\bibitem[{Perchtold-Stefan et~al.(2024)Perchtold-Stefan, Kapoor, Kaufman, Mahadeshwar, and Fernandes}]{perchtold2024development}
Corinna Perchtold-Stefan, Hansika Kapoor, James~C Kaufman, Hreem Mahadeshwar, and Alison Fernandes. 2024.
\newblock Development and neuronal validation of the dark creativity deception battery (dcdb).

\bibitem[{Perchtold-Stefan et~al.(2023)Perchtold-Stefan, Rominger, Papousek, and Fink}]{perchtold2023functional}
Corinna~M Perchtold-Stefan, Christian Rominger, Ilona Papousek, and Andreas Fink. 2023.
\newblock Functional eeg alpha activation patterns during malevolent creativity.
\newblock \emph{Neuroscience}, 522:98--108.

\bibitem[{Rafailov et~al.(2023)Rafailov, Sharma, Mitchell, Manning, Ermon, and Finn}]{rafailov2023direct}
Rafael Rafailov, Archit Sharma, Eric Mitchell, Christopher~D Manning, Stefano Ermon, and Chelsea Finn. 2023.
\newblock Direct preference optimization: Your language model is secretly a reward model.
\newblock \emph{Advances in Neural Information Processing Systems}, 36:53728--53741.

\bibitem[{Rajbhandari et~al.(2020)Rajbhandari, Rasley, Ruwase, and He}]{rajbhandari2020zero}
Samyam Rajbhandari, Jeff Rasley, Olatunji Ruwase, and Yuxiong He. 2020.
\newblock Zero: Memory optimizations toward training trillion parameter models.
\newblock In \emph{SC20: International Conference for High Performance Computing, Networking, Storage and Analysis}, pages 1--16. IEEE.

\bibitem[{Raz et~al.(2024)Raz, Luchini, Beaty, and Kenett}]{raz2024bridging}
Tuval Raz, Simone Luchini, Roger Beaty, and Yoed Kenett. 2024.
\newblock Bridging the measurement gap: A large language model method of assessing open-ended question complexity.
\newblock In \emph{Proceedings of the Annual Meeting of the Cognitive Science Society}, volume~46.

\bibitem[{Runco and Jaeger(2012)}]{runco2012standard}
Mark~A Runco and Garrett~J Jaeger. 2012.
\newblock The standard definition of creativity.
\newblock \emph{Creativity research journal}, 24(1):92--96.

\bibitem[{S{\ae}b{\o} and Brovold(2024)}]{saebo2024stochastics}
Solve S{\ae}b{\o} and Helge Brovold. 2024.
\newblock On the stochastics of human and artificial creativity.
\newblock \emph{arXiv preprint arXiv:2403.06996}.

\bibitem[{Saito et~al.(2023)Saito, Wachi, Wataoka, and Akimoto}]{saito2023verbositybiaspreferencelabeling}
Keita Saito, Akifumi Wachi, Koki Wataoka, and Youhei Akimoto. 2023.
\newblock \href {https://arxiv.org/abs/2310.10076} {Verbosity bias in preference labeling by large language models}.
\newblock \emph{Preprint}, arXiv:2310.10076.

\bibitem[{Saretzki et~al.(2024)Saretzki, Andrae, Forthmann, and Benedek}]{saretzki2024investigation}
Janika Saretzki, Rosalie Andrae, Boris Forthmann, and Mathias Benedek. 2024.
\newblock Investigation of response aggregation methods in divergent thinking assessments.
\newblock \emph{The Journal of Creative Behavior}.

\bibitem[{Schroeders et~al.(2016)Schroeders, Wilhelm, and Olaru}]{schroeders2016meta}
Ulrich Schroeders, Oliver Wilhelm, and Gabriel Olaru. 2016.
\newblock Meta-heuristics in short scale construction: Ant colony optimization and genetic algorithm.
\newblock \emph{PloS one}, 11(11):e0167110.

\bibitem[{Shypula et~al.(2025)Shypula, Li, Zhang, Padmakumar, Yin, and Bastani}]{shypula2025evaluating}
Alexander Shypula, Shuo Li, Botong Zhang, Vishakh Padmakumar, Kayo Yin, and Osbert Bastani. 2025.
\newblock Evaluating the diversity and quality of llm generated content.
\newblock \emph{arXiv preprint arXiv:2504.12522}.

\bibitem[{Silvia(2011)}]{silvia2011subjective}
Paul~J Silvia. 2011.
\newblock Subjective scoring of divergent thinking: Examining the reliability of unusual uses, instances, and consequences tasks.
\newblock \emph{Thinking Skills and Creativity}, 6(1):24--30.

\bibitem[{Simonton(2012)}]{simonton2012taking}
Dean~Keith Simonton. 2012.
\newblock Taking the us patent office criteria seriously: A quantitative three-criterion creativity definition and its implications.
\newblock \emph{Creativity research journal}, 24(2-3):97--106.

\bibitem[{Simonton(2018)}]{simonton2018defining}
Dean~Keith Simonton. 2018.
\newblock Defining creativity: Don't we also need to define what is not creative?
\newblock \emph{The Journal of Creative Behavior}, 52(1):80--90.

\bibitem[{Stein(2014)}]{stein2014stimulating}
Morris~I Stein. 2014.
\newblock \emph{Stimulating creativity: Individual procedures}.
\newblock Academic Press.

\bibitem[{Stevenson et~al.(2022)Stevenson, Smal, Baas, Grasman, and van~der Maas}]{stevenson2022putting}
Claire~E. Stevenson, Iris Smal, Matthijs Baas, Raoul Grasman, and Han L.~J. van~der Maas. 2022.
\newblock \href {https://arxiv.org/abs/2206.08932} {Putting gpt-3's creativity to the (alternative uses) test}.
\newblock In \emph{ICCC}.

\bibitem[{Sturua et~al.(2024)Sturua, Mohr, Akram, Günther, Wang, Krimmel, Wang, Mastrapas, Koukounas, Koukounas, Wang, and Xiao}]{sturua2024jinaembeddingsv3multilingualembeddingstask}
Saba Sturua, Isabelle Mohr, Mohammad~Kalim Akram, Michael Günther, Bo~Wang, Markus Krimmel, Feng Wang, Georgios Mastrapas, Andreas Koukounas, Andreas Koukounas, Nan Wang, and Han Xiao. 2024.
\newblock \href {https://arxiv.org/abs/2409.10173} {jina-embeddings-v3: Multilingual embeddings with task lora}.
\newblock \emph{Preprint}, arXiv:2409.10173.

\bibitem[{Summers-Stay et~al.(2023)Summers-Stay, Lukin, and Voss}]{SummersStay2023BrainstormTS}
Douglas Summers-Stay, Stephanie~M. Lukin, and Clare~R. Voss. 2023.
\newblock \href {https://api.semanticscholar.org/CorpusID:259305709} {Brainstorm, then select: a generative language model improves its creativity score}.

\bibitem[{Sun et~al.(2025)Sun, Pei, Choi, and Jurgens}]{sun2025sociodemographic}
Huaman Sun, Jiaxin Pei, Minje Choi, and David Jurgens. 2025.
\newblock Sociodemographic prompting is not yet an effective approach for simulating subjective judgments with llms.
\newblock In \emph{Proceedings of the 2025 Conference of the Nations of the Americas Chapter of the Association for Computational Linguistics: Human Language Technologies (Volume 2: Short Papers)}, pages 845--854.

\bibitem[{Team et~al.(2023)Team, Anil, Borgeaud, Alayrac, Yu, Soricut, Schalkwyk, Dai, Hauth, Millican et~al.}]{team2023gemini}
Gemini Team, Rohan Anil, Sebastian Borgeaud, Jean-Baptiste Alayrac, Jiahui Yu, Radu Soricut, Johan Schalkwyk, Andrew~M Dai, Anja Hauth, Katie Millican, and 1 others. 2023.
\newblock Gemini: a family of highly capable multimodal models.
\newblock \emph{arXiv preprint arXiv:2312.11805}.

\bibitem[{Tian et~al.(2024)Tian, Huang, Liu, Jiang, Spangher, Chen, May, and Peng}]{tian2024large}
Yufei Tian, Tenghao Huang, Miri Liu, Derek Jiang, Alexander Spangher, Muhao Chen, Jonathan May, and Nanyun Peng. 2024.
\newblock Are large language models capable of generating human-level narratives?
\newblock \emph{arXiv preprint arXiv:2407.13248}.

\bibitem[{Tian et~al.(2023)Tian, Ravichander, Qin, Bras, Marjieh, Peng, Choi, Griffiths, and Brahman}]{tian2023macgyver}
Yufei Tian, Abhilasha Ravichander, Lianhui Qin, Ronan~Le Bras, Raja Marjieh, Nanyun Peng, Yejin Choi, Thomas~L Griffiths, and Faeze Brahman. 2023.
\newblock Macgyver: Are large language models creative problem solvers?
\newblock \emph{arXiv preprint arXiv:2311.09682}.

\bibitem[{Wang et~al.(2024{\natexlab{a}})Wang, Zheng, Chen, Xi, Shen, Zhou, Yan, Gui, Zhang, and Huang}]{wang2024reward}
Binghai Wang, Rui Zheng, Lu~Chen, Zhiheng Xi, Wei Shen, Yuhao Zhou, Dong Yan, Tao Gui, Qi~Zhang, and Xuan-Jing Huang. 2024{\natexlab{a}}.
\newblock Reward modeling requires automatic adjustment based on data quality.
\newblock In \emph{Findings of the Association for Computational Linguistics: EMNLP 2024}, pages 4041--4064.

\bibitem[{Wang et~al.(2024{\natexlab{b}})Wang, Chen, Jia, Wang, Fang, Wang, Gao, Xie, Xu, Dai et~al.}]{wang2024weaver}
Tiannan Wang, Jiamin Chen, Qingrui Jia, Shuai Wang, Ruoyu Fang, Huilin Wang, Zhaowei Gao, Chunzhao Xie, Chuou Xu, Jihong Dai, and 1 others. 2024{\natexlab{b}}.
\newblock Weaver: Foundation models for creative writing.
\newblock \emph{arXiv preprint arXiv:2401.17268}.

\bibitem[{Wei et~al.(2022)Wei, Tay, Bommasani, Raffel, Zoph, Borgeaud, Yogatama, Bosma, Zhou, Metzler et~al.}]{wei2022emergent}
Jason Wei, Yi~Tay, Rishi Bommasani, Colin Raffel, Barret Zoph, Sebastian Borgeaud, Dani Yogatama, Maarten Bosma, Denny Zhou, Donald Metzler, and 1 others. 2022.
\newblock Emergent abilities of large language models.
\newblock \emph{arXiv preprint arXiv:2206.07682}.

\bibitem[{Weiss et~al.(2024)Weiss, Goecke, and Wilhelm}]{weiss2024much}
Selina Weiss, Benjamin Goecke, and Oliver Wilhelm. 2024.
\newblock How much retrieval ability is in originality?
\newblock \emph{The Journal of Creative Behavior}, 58(3):370--387.

\bibitem[{Weiss et~al.(2023)Weiss, Olderbak, and Wilhelm}]{weiss2023conceptualizing}
Selina Weiss, Sally Olderbak, and Oliver Wilhelm. 2023.
\newblock Conceptualizing and measuring ability emotional creativity.
\newblock \emph{Psychology of Aesthetics, Creativity, and the Arts}.

\bibitem[{Wenger and Kenett(2025)}]{wenger2025we}
Emily Wenger and Yoed Kenett. 2025.
\newblock We're different, we're the same: Creative homogeneity across llms.
\newblock \emph{arXiv preprint arXiv:2501.19361}.

\bibitem[{West and Potts(2025)}]{west2025basemodelsbeataligned}
Peter West and Christopher Potts. 2025.
\newblock \href {https://arxiv.org/abs/2505.00047} {Base models beat aligned models at randomness and creativity}.
\newblock \emph{Preprint}, arXiv:2505.00047.

\bibitem[{Wilkinson(2023)}]{Wilkinson_2023}
Alissa Wilkinson. 2023.
\newblock \href {https://www.vox.com/culture/23696617/writers-strike-wga-2023-explained-residuals-streaming-ai} {Hollywood’s writers are on strike. here’s why that matters.}

\bibitem[{Wong et~al.(2024)Wong, Orlovskiy, Luo, Seshia, and Gonzalez}]{wong2024simplestrat}
Justin Wong, Yury Orlovskiy, Michael Luo, Sanjit~A Seshia, and Joseph~E Gonzalez. 2024.
\newblock Simplestrat: Diversifying language model generation with stratification.
\newblock \emph{arXiv preprint arXiv:2410.09038}.

\bibitem[{Xu et~al.(2024)Xu, Jojic, Rao, Brockett, and Dolan}]{xu2024echoes}
Weijia Xu, Nebojsa Jojic, Sudha Rao, Chris Brockett, and Bill Dolan. 2024.
\newblock Echoes in ai: Quantifying lack of plot diversity in llm outputs.
\newblock \emph{arXiv preprint arXiv:2501.00273}.

\bibitem[{Yeh et~al.(2024)Yeh, Tao, Wang, Du, and Li}]{yeh2024reliable}
Min-Hsuan Yeh, Leitian Tao, Jeffrey Wang, Xuefeng Du, and Yixuan Li. 2024.
\newblock How reliable is human feedback for aligning large language models?
\newblock \emph{arXiv preprint arXiv:2410.01957}.

\bibitem[{Yu et~al.(2024)Yu, Krebs, Beeman, and Lai}]{yu2024exploring}
Yuhua Yu, Lindsay Krebs, Mark Beeman, and Vicky~T Lai. 2024.
\newblock Exploring how generating metaphor via insight versus analysis affects metaphor quality and learning outcomes.
\newblock \emph{Cognitive science}, 48(8):e13488.

\bibitem[{Zhang et~al.(2025)Zhang, Diddee, Holm, Liu, Liu, Samuel, Wang, and Ippolito}]{zhang2025noveltybench}
Yiming Zhang, Harshita Diddee, Susan Holm, Hanchen Liu, Xinyue Liu, Vinay Samuel, Barry Wang, and Daphne Ippolito. 2025.
\newblock Noveltybench: Evaluating creativity and diversity in language models.
\newblock \emph{arXiv preprint arXiv:2504.05228}.

\bibitem[{Zhang et~al.(2024)Zhang, Schwarzschild, Carlini, Kolter, and Ippolito}]{zhang2024forcing}
Yiming Zhang, Avi Schwarzschild, Nicholas Carlini, Zico Kolter, and Daphne Ippolito. 2024.
\newblock Forcing diffuse distributions out of language models.
\newblock \emph{arXiv preprint arXiv:2404.10859}.

\bibitem[{Zhao et~al.(2025)Zhao, Zhou, Li, Tang, Wang, Hou, Min, Zhang, Zhang, Dong, Du, Yang, Chen, Chen, Jiang, Ren, Li, Tang, Liu, Liu, Nie, and Wen}]{zhao2025surveylargelanguagemodels}
Wayne~Xin Zhao, Kun Zhou, Junyi Li, Tianyi Tang, Xiaolei Wang, Yupeng Hou, Yingqian Min, Beichen Zhang, Junjie Zhang, Zican Dong, Yifan Du, Chen Yang, Yushuo Chen, Zhipeng Chen, Jinhao Jiang, Ruiyang Ren, Yifan Li, Xinyu Tang, Zikang Liu, and 3 others. 2025.
\newblock \href {https://arxiv.org/abs/2303.18223} {A survey of large language models}.
\newblock \emph{Preprint}, arXiv:2303.18223.

\bibitem[{Zhao et~al.(2024)Zhao, Zhang, Li, Huang, Guo, Peng, Hao, Wen, Hu, Du et~al.}]{zhao2024assessing}
Yunpu Zhao, Rui Zhang, Wenyi Li, Di~Huang, Jiaming Guo, Shaohui Peng, Yifan Hao, Yuanbo Wen, Xing Hu, Zidong Du, and 1 others. 2024.
\newblock Assessing and understanding creativity in large language models.
\newblock \emph{arXiv preprint arXiv:2401.12491}.

\bibitem[{Zhou et~al.(2025)Zhou, Chen, Suresh, Narad, Rogers, Jain, Nowak, Mankoff, and Zhang}]{zhou2025bridging}
Kuan~Lok Zhou, Jiayi Chen, Siddharth Suresh, Reuben Narad, Timothy~T Rogers, Lalit~K Jain, Robert~D Nowak, Bob Mankoff, and Jifan Zhang. 2025.
\newblock Bridging the creativity understanding gap: Small-scale human alignment enables expert-level humor ranking in llms.
\newblock \emph{arXiv preprint arXiv:2502.20356}.

\bibitem[{Zieli{\'n}ska et~al.(2023)Zieli{\'n}ska, Organisciak, Dumas, and Karwowski}]{zielinska2023lost}
Aleksandra Zieli{\'n}ska, Peter Organisciak, Denis Dumas, and Maciej Karwowski. 2023.
\newblock Lost in translation? not for large language models: Automated divergent thinking scoring performance translates to non-english contexts.
\newblock \emph{Thinking Skills and Creativity}, 50:101414.

\end{thebibliography}
